\documentclass[11pt]{amsart}
\usepackage[centertags]{amsmath}
\usepackage{amsfonts}
\usepackage{amssymb}
\usepackage{amsthm}
\usepackage[english]{babel}
\usepackage[latin2]{inputenc}
\usepackage{graphicx}

\title[Controlled hierarchical filtering]{Controlled hierarchical filtering: Model of neocortical sensory processing}

\author[A. L\H{o}rincz]{Andr\'{a}s L\H{o}rincz}
       \thanks{\newline Department of Information Systems \\ E\"otv\"os Lor\'and University \\
              P\'azm\'any P\'eter s\'et\'any 1/C \\
              Budapest, Hungary H-1117 \\
              email: lorincz@inf.elte.hu}

\begin{document}
\maketitle

\begin{abstract}
A model of sensory information processing is presented. The model assumes that learning of internal (hidden) generative
models, which can predict the future and evaluate the precision of that prediction, is of central importance for
information extraction. Furthermore, the model makes a bridge to goal-oriented systems and builds upon the structural
similarity between the architecture of a robust controller and that of the hippocampal entorhinal loop. This generative
control architecture is mapped to the neocortex and to the hippocampal entorhinal loop. Implicit memory phenomena;
priming and prototype learning are emerging features of the model. Mathematical theorems ensure stability and
attractive learning properties of the architecture. Connections to reinforcement learning are also established: both
the control network, and the network with a hidden model converge to (near) optimal policy under suitable conditions.
Falsifying predictions, including the role of the feedback connections between neocortical areas are made.
\end{abstract}

\newpage
\tableofcontents
\newpage

\section{Introduction}
\label{s:intro}

Our thinking is best expressed by the words of Albert Szent-Gy\"orgyi \cite{szentgyorgyi51}, the famous Hungarian Nobel
Laureate: `There is no real difference between structure and function; they are the two sides of the same coin. If
structure does not tell us anything about function, it means we have not looked at it correctly.'

Here, a general framework is searched for that explains
information processing function of the brain, extends to the
goal-oriented nature of information processing and explains the
structure given these functions. For the sake of clarity, first
our assumptions about the function shall be stated. Built on the
assumptions, we shall work on `deriving' the building blocks of
structure. Finally, these blocks will be mapped to the substrate.

\subsection{Starting assumptions} \label{ss:assume}

The model is built on a few  interlinked `axioms':
\begin{enumerate}
    \item The brain interacts with the environment, signals are
    received and responses are generated.
    \item Signal detection corresponds to \textit{filtering} of
    all possible detectable phenomena.
    \item Response generation influences the environment, influence on the environment is goal oriented and aims
    to \textit{control} environmental parameters.
    \item Control of the environment is subject to optimization.
\end{enumerate}

\subsection{`Guiding principles'} \label{ss:guide}

Our starting assumptions are constrained as follows:

\subsubsection*{The homunculus fallacy should be solved} \label{ss:hom}
Our thoughts are grounded  on the hypothesis that representations do exist in the brain (see e.g. the debates about the
Representational Theory of Mind and its modern extension, the Computational Theory of Mind
\cite{fodor81representations,churchland89onthenature}, but also \cite{dennett87theintentional}). The use of
representations can hardly be avoided in any computational modelling. Generally speaking, the processing of signals
that may convey information can be considered as a transformation into another form that still carries the whole amount
or just a piece of the original information. The environment feeds the system with some inputs and the system output
represents (a part of) the environment. Whilst most models address the problem of coding inputs and making efficient
internal representation, we are more concerned about the fundamental problem of making sense of these representations.
In our view, the central issue of making sense or \textit{meaning} is to provide answers to questions like `what does
it mean?' in terms of our past experiences, or `how are they related?' in terms of known facts. In other words, making
sense is inherently related to declarative memory. As a consequence, the homunculus fallacy (see e.g.,
\cite{searle92rediscovery}) --- that the internal representation is meaningless without an interpreter --- is of
central importance. This fallacy claims that all levels of abstraction require at least one further level to become the
corresponding interpreter. Unfortunately, the interpretation --- according to the fallacy --- is just a new
transformation and we are trapped in an endless regression.\footnote{We note there can be more than one route to
resolve the fallacy (see, e.g., \cite{dennett91consiousness}). Along the line of the classical black box modeling the
fallacy does not arise at all, but \textit{meaningful labeling} of blocks of the model can be questioned.}

\subsubsection{Constraint of reconstruction}

Our standpoint is that the paradox  stems from vaguely described procedure of `making sense'. The fallacy arises by
saying that the internal representation should make sense. One can turn the fallacy upside down by changing the roles
\cite{lorincz97towards}: Not the internal representation but the \textit{input} should make sense. Our proposal is that
the \emph{input makes sense} if the same (or similar) inputs have been experienced before and if the input can be
derived or regenerated by means of the internal representation \cite{lorincz97towards,lorincz01recognition}. All in
all, the goal is to turn the infinite regression into a reconstructing loop structure and shortcut the fallacy.
According to this approach the internal representation interprets the input by (re)constructing it. This function is
more than \textit{mirroring} the environment. Interpretation based reconstruction can fill in missing parts of
spatio-temporal patterns\footnote{Here, `spatial' means information sets processed almost simultaneously. For example,
almost simultaneous retinotopic information or, information about different audio frequencies in the auditory cortex,
or both of these, etc., are called spatial components.}, which includes the capacity of prediction.

\subsubsection{Constraints of control and optimization} \label{ss:RL}

Interpretation is goal oriented and forms a delicate

`perception-action loop': The actual goal requires (1) the sensing
of environmental parameters, (2) the ability to influence, i.e, to
\textit{control} those parameters, (3) the sensing of
consequences of control and so on. In turn, perceptual information is transformed, transformation is controlled and the control of transformation is subject to the actual goal. Because transformation depends on the actual goal, control typically acts on a partially observed environment; information is filtered.

Sensing and control of the environment consumes energy,  which
should be minimized. Minimization of energy consumption is a
long-term task: short term saving at the cost of large long-term
spending needs to be avoided and long-term cumulated cost is to be
minimized. Such optimization problems are formulated within the
framework of reinforcement learning (RL)\footnote{For an excellent
introductory materials on RL, see, e.g.,
\cite{Sutton98Reinforcement}.}. It is then necessary to consider
control concepts subject to principles developed in RL in
partially observed environments. This challenging
\cite{aaai98pomdp} and generally computationally intractable
problem \cite{littman95learning,littman96algorithms} should be
addressed by the model.

\subsubsection{Architectural constraints on the neural level} \label{ss:Hebb}

There are constraints on the building blocks of the filtering and
controlling system, such as
\begin{enumerate}
    \item Locality: The architecture is made of connected simple computational
    units. Connected units are \textit{neighbors}. Computations of any unit
    are based on information received from its neighbors.
    \item Connections are directed and serve as filters. Connections possess tunable \textit{filtering
    strength} or weights. Tuning of connections is
    subject to \textit{Hebbian-learning}: signals of the two computational units
    at the two ends of a directed connection determine the adaptation
    of the weight of the connection.
    \item Locality and Hebbian-learning concern all functions, including
    sensing, control and optimization.
\end{enumerate}

\subsubsection{Anatomical constraints} \label{ss:neuro}

The architecture should  match known architectural properties of
sensory processing areas of the neocortex up to the top, the
hippocampus (HC) and its surrounding, the hippocampal formation.

\subsection{Origins of the model}\label{ss:origins}

Ever since the discovery of  the central role of the hippocampus
and its adjacent areas in memory formation
\cite{sidman68some,milner72disorders}, numerous studies and models
dealt with the properties and the possible functions of the
hippocampus and its environment. The number of new experimental
findings is increasing and highlight the complexity of the
behavior of memory. Although views are strikingly different, they
seem to have their own, experimentally supported merits. The
interested reader is referred to the literature for excellent
reviews on the hippocampus written, e.g., by Squire
\cite{squire92memory}, Hasselmo and McClelland
\cite{hasselmo99neural}, Redish \cite{redish99beyond} and O'Reilly
and Rudy \cite{oreilly99conjunctive}. The majority of the models
have been developed to describe one part (mainly the CA3 field) of
the hippocampus (see, e.g.
\cite{levy96sequence,kali00involvement}). Attempts have been made
to develop an integrating model of the HC
\cite{rolls89functions,hasselmo96encoding,lisman99relating,eichenbaum00cortical,hasselmo02proposed}.
See also works collected by Gluck \cite{hippocampus96gluck}. It is
known though, that hippocampus is deeply embedded in the
neocortical information flow through the entorhinal cortex (EC).

This fact explains the emergence of a few EC-HC models like
\cite{mcclelland95why,myers95dissociation,lorincz98forming,rolls00hippocampo,lorincz02mystery,oreilly02hippocampal}.
For example, McClelland et al. \cite{mcclelland95why} emphasize
the necessity of a dual system for the seemingly contradictory
tasks of learning of specific properties and allowing for
generalization.

The controlled hierarchical  filtering (CHF) model that we present
here, has its origin in the old standing proposal that the
hippocampus and/or its environment serve as a `comparator'
\cite{grastyan59hippocampal,sokolov63higher,vinogradova75functional}.
More recent works about this subject try to provide a neuro-psychological account of anxiety and consciousness
\cite{gray82neuropsychology,gray82precis,gray86comparator,gray95contents}.
Other models use somewhat different nomenclature, e.g., the focus
is placed on match/mismatch detection
\cite{ranck73studies,okeefe78hippocampus,grossberg82processing}.
Match/mismatch detection is closely related to familarity/novelty
detection, another direction of theoretical efforts to describe
medial temporal lobe areas
\cite{otto92neuronal,rolls93responses,wiebe88dynamic}. The form of
novelty is probably polymorphous and there is increasing evidence
that different brain areas share the task of recognizing different
aspects of novelty \textit{within} the same scene
\cite{wan99different}. It seems that the encoding of novelty is
distributed, which is a crucial point of the CHF model.

The CHF architecture is an extension of our previous works
\cite{lorincz98forming,lorincz00parahippocampal,lorincz02mystery}.
It may be worth noting that two falsifying predictions of that
model, (a) large and tunable temporal delaying capabilities of
neurons of the dentate gyrus and (b) persistent activities at deep
layers of the entorhinal cortex have been reinforced recently by
Henze et al., \cite{henze02single} and by Egorov et al.
\cite{egorov02graded}, respectively.

The paper is constructed as follows.  First (in Section
\ref{s:model}), terminology is provided and basic concepts are
defined. The CHF model is detailed here. Section
\ref{s:mapping} deals with the mapping of the control and

reconstruction architectures to the entorhinal-hippocampal loop
and to neocortical areas. Section \ref{s:disc} discusses relations
to other computational models, e.g.,

\cite{gluck93hippocampal,stainvas00improving,rao97dynamic,rao99predictive,bousquet99is}. This section treats learning and
stability properties of the architecture, connections between RL
and the model architecture, connections to RL and to partially
observed RL problems, the emerging neurobiological features,
missing links and some conjectures of the model. The paper is
finished by an Appendix containing some of the mathematical
details. Other mathematical details can be found in the cited
references, including the technicalities, which are made available
as technical reports through http://arxiv.org.

\section{Model construction}
\label{s:model}

\begin{figure}[h]
\centering
 \includegraphics[width=76mm]{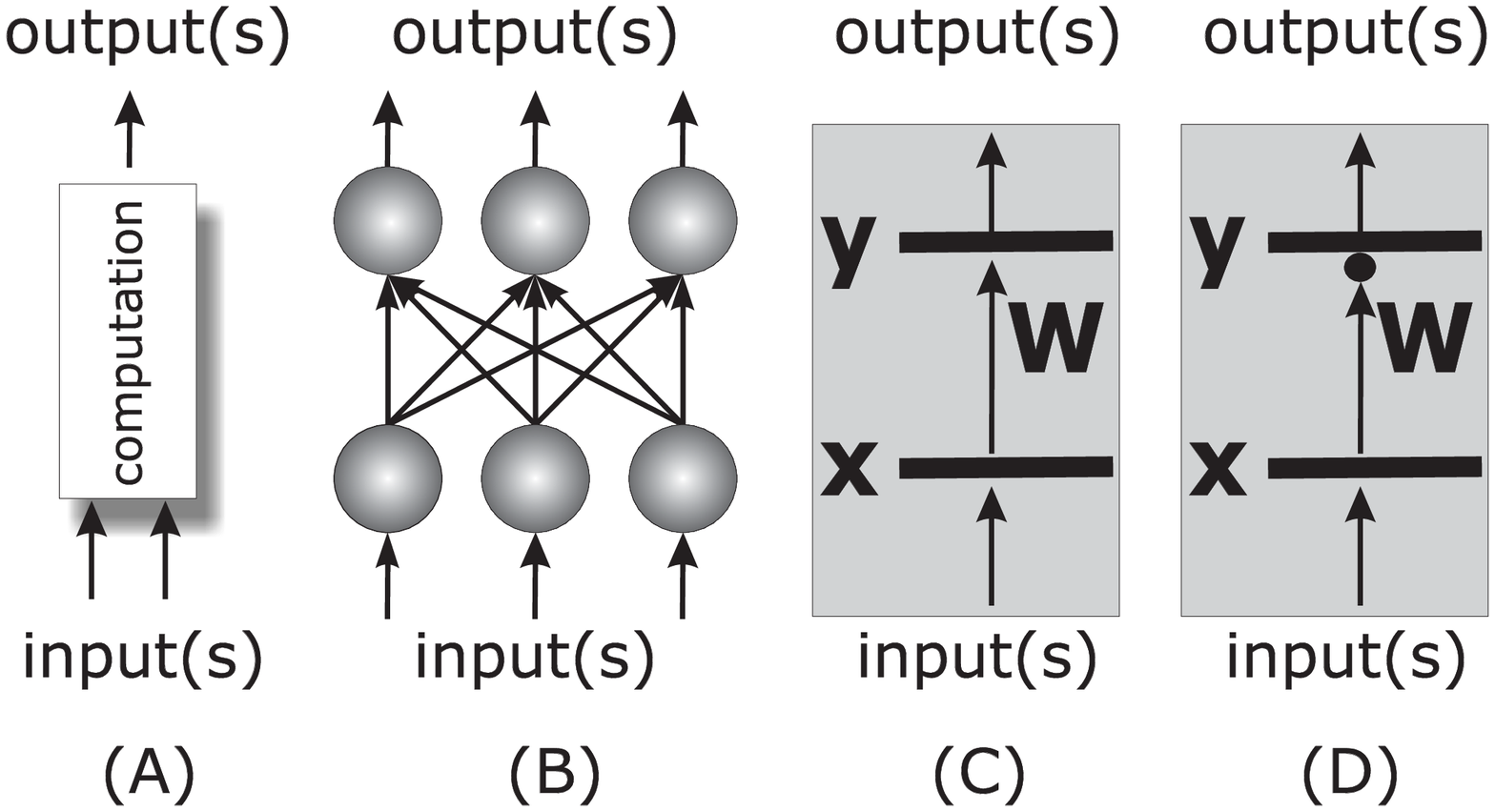}
 \caption{\textbf{Notations}\newline
  \textbf{A:} Control representation of input output systems.
  Computations are performed in the box. \newline \textbf{B:} Neural
  network representation of computations: inputs are received by
  input neurons and are (non-linearly) transformed by connections
  and the output neurons, which provide the outputs.
  An output neuron could be an input neuron of the next processing
  stage. \newline \textbf{C:} Linear neural transformation. Input: $\mathbf{x}$,
  transformation $\mathbf{W}$, output: $\mathbf{y}$,
  $\mathbf{y}=\mathbf{W}\mathbf{x}$. \newline \textbf{D:} Non-linear neural transformation.
  $\mathbf{y}=f(\mathbf{W}\mathbf{x})$. Graphical form: arrow with a circle. More than one
  transformation may exist between layers. Recurrent network
  is a neural layer with a transformation that targets the same
  layer.
  \newline \textit{Terminology in the context of neurobiology:} \newline Layer corresponds to a given
  area sometimes called field or subfield, such as the CA3 and CA1 regions of
  the hippocampus, or the different layers of the neocortex.
  Transformations may correspond to (i) excitatory synapses
  connecting layers or targeting neurons of the same layer, such
  as the \textit{recurrent collaterals} and the \textit{associative connections}
  of the CA3 subfield of the hippocampus and the intra-layer excitatory connections
  of layers II and III of the neocortex or (ii) inhibitory synapses
  between layers or within layers, such as the rich interneural networks
  in the hippocampus.}
\label{f:notat}
\end{figure}

Notations of the control field and notations of neuron networks
differ. In control theory, input-output systems are considered. In
graphical form, box denotes the system and arrows  denote the
system's input(s)  and output(s). Processing occurs in the box
(Fig.~\ref{f:notat}(A)). On the contrary, artificial neural
networks consist of computational units, the putative analogs of
real neurons. The units, also called neurons, receive (provide)
inputs (outputs) through the connection structure, and this
internal functioning is drawn explicitly. Neurons execute simple
computations, like summing up inputs, thresholding and alike. The
main part of the neural network performs distributed computation
using the connection structure performing (non-linear) filtering.
This distributed filter system, which may connect all neurons, is
called the connection system, weights, or synapses. In a neural
network architecture different neural layers are distinguished.
Connections between these layers are explicitly drawn in most
cases (Fig.~\ref{f:notat}(B)). Computations of neural networks
between their inputs and outputs can be given in the following
condensed form:
\begin{equation}\label{eq:neur}
\mathbf{y} = f(\mathbf{Wx})
\end{equation}
where input and output  are denoted by $\mathbf{x} \in
\mathbb{R}^n$ and $\mathbf{y} \in \mathbb{R}^m$, respectively,
linear transformation from $\mathbb{R}^n$ to $\mathbb{R}^m$ is
represented by matrix $\mathbf{W} \in \mathbb{R}^{m \times n}$,
the connections, and function $f$ denotes components-wise
non-linearity. If this function is the identity function, then we
have a linear network. Here, a simplified notation will be used:
neural layers will be denoted by horizontal thick lines. Any
particular set of connections between two layers will be
represented by a single arrow. A feedforward linear network is
depicted in Fig.~\ref{f:notat}(C). The graphical form of a network
with component-wise non-linearity is shown in
Fig.~\ref{f:notat}(D). Different transformations may exist between
two layers. \textit{Recurrent connections} (also called `recurrent
collaterals') target the same layer where they originate from.

\subsection{The control model} \label{ss:control}

\begin{figure}
\centering
 \includegraphics[width=50mm]{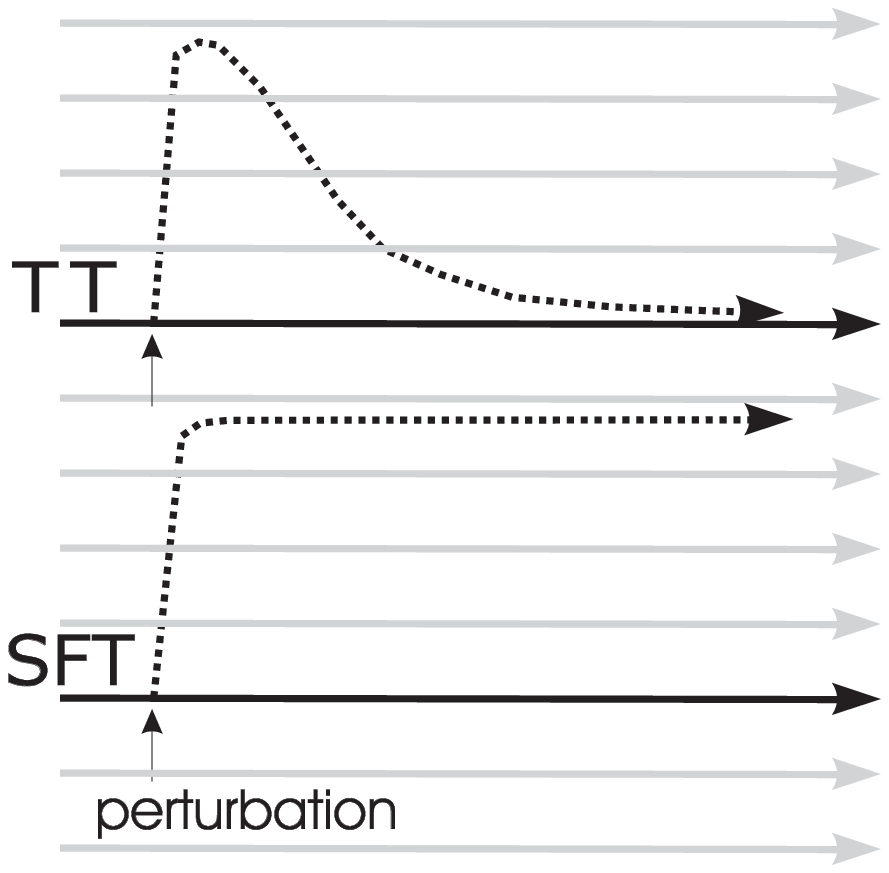}
  \caption{\textbf{Trajectory tracking (TT) versus speed field tracking (SFT)}\newline
  \textbf{TT:} Horizontal lines represent different nearby trajectories.
  Black line: initial trajectory. Upon perturbations the system (the `plant') should return
  to the original trajectory.
  \newline \textbf{SFT:} Horizontal lines represent a small part of the speed-field to be tracked.
  Black line: initial speed trajectory. Upon perturbation, the plant
  adjusts its speed to the speed of the actual neighborhood.}\label{f:SFT}
\end{figure}

Our control problem  is formulated in terms of state dependent
directions pointing towards target positions. A mapping which
renders direction (change of state, or change of state per unit
time, i.e., velocity) to each state is called speed-field. A
particular speed-field is given, for example, by the difference
vectors between the target state and all other states. An
important feature of speed-field is that motion is not specified
in time. The control task is defined as moving according to the
speed-field at each state. This control task is called speed-field
tracking (SFT). For a review on SFT, see, e.g., \cite{HwaAh92}.
SFT formulation is flexible because different fields can be
designed for the same task and, also, it allows motions to speed
up or to slow down simply by scaling of the speed-field.
Speed-field can be seen as a local tool for path planning (see,
e.g., \cite{FoRoSzLo96IJNS} and references therein). The control
task of path (also called trajectory) tracking is, however,
different. The difference between SFT and ordinary trajectory
tracking is shown in Fig.~\ref{f:SFT}. One might say that SFT is
less stringent, less precise and puts more emphasis on the global
goal than on the local perturbations.

The dynamic equation  of a system is a (possibly continuous) set of differential equations. This set of equations
determines the change of state per unit time given the external forces acting upon the system, including the control
action. Inverse dynamics works in the opposite way: given the state and (desired) change of state, inverse dynamics
provides the control vector. If the inverse dynamics is perfect then inserting the control vector into the dynamic
equation the desired change of state is achieved. The controller, in turn, maps state and speed to control action.

Inverse dynamics,  however, changes if kinematic
parameters (such as dimensions; length, width, etc.), or
parameters of the dynamics (e.g., weight, flexibility and so on) change: Inverse
dynamics is almost never perfect. Moreover, it is well known, that
approximate inverse dynamics can give rise to instabilities
\cite{isidori89nonlinear}. In turn, a robust extension of
speed-field tracking is needed. Such robust control architecture
is described here.

\begin{figure}
\centering
 \includegraphics[width=76mm]{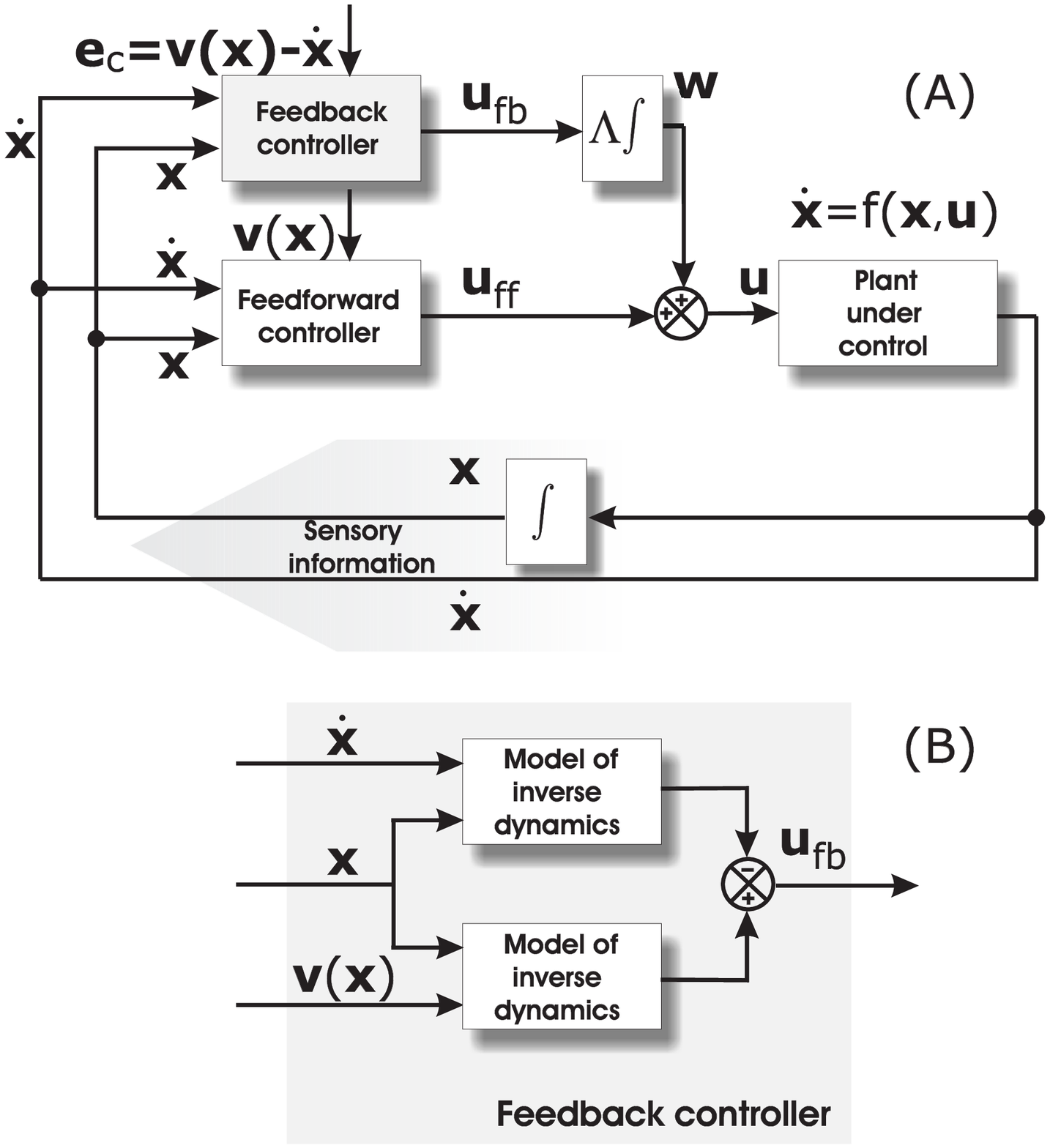}
  \caption{\textbf{Robust controller for speed-field tracking tasks}\newline
  \textbf{A:} The model of the inverse dynamics is inputted by the actual state $\mathbf{x}$, the
  momentum $\mathbf{\dot{x}}= \frac{d \mathbf{x}}{d t}$ , and
  the desired momentum at the actual state $\mathbf{v}(\mathbf{x})$.
  The output of the model is the \textit{feedforward} control vector $\mathbf{u}_{ff}=\mathbf{u}_{ff}(\mathbf{x},\mathbf{\dot{x}},\mathbf{v}(\mathbf{x}))$.
  The feedforward control vector
  may need corrections. The \textit{feedback} control vector, which is inputted by the state,
  the momentum and the desired compensation
  $\mathbf{e}_c=\mathbf{v}(\mathbf{x})-\mathbf{\dot{x}}$ serves this
  purpose.   The output of the feedback controller is integrated by time, it is multiplied by the gain factor
  $\Lambda$ and the result is added to the feedforward control vector to form the approximate control
  vector $\mathbf{u}$.
  \newline \textbf{B:} The feedback controller is composed of two simplified models
  of the inverse dynamics. Their effect cancels and, in turn, feedback control action disappears when the
  feedforward controller is perfect, i.e., when control vector $\mathbf{u}_{ff}$ produces the
  desired momentum: $\mathbf{\dot{x}}=\mathbf{v}(\mathbf{x})$.
  These models have two arguments, the state and the momentum.
  The first model is inputted by the actual state and the desired momentum. The output
  of the model makes a positive contribution. The second model uses
  the actual state and the actual momentum. The difference of the two
  outputs,
  $\mathbf{u}_{fb}=\mathbf{\hat{\Phi}}(\mathbf{x},\mathbf{v}(\mathbf{x}))-\mathbf{\hat{\Phi}}(\mathbf{x},\mathbf{\dot{x}})$,
  is the feedback control vector.}\label{f:robust_controller}
\end{figure}

Let $\mathbf{x}$ and $\mathbf{\dot{x}}$, where `dot' denotes temporal derivation, represent the state and the change of
the state per unit time (the momentum) of the plant, respectively. Let
\begin{equation}\label{eq:plant0}
\mathbf{\dot{x}} = \mathbf{f}(\mathbf{x},\mathbf{u})
\end{equation}
denote the dynamics  of the plant, i.e., in state $\mathbf{x} \in \mathbb{R}^n$ and under control $\mathbf{u} \in
\mathbb{R}^p$, the momentum of the plant becomes $\mathbf{\dot{x}} \in \mathbb{R}^n$ described by the nonlinear
function $\mathbf{f}:\mathbb{R}^{n \times p} \rightarrow \mathbb{R}^n $. Let $\mathbf{v}(\mathbf{x}) \in \mathbb{R}^n$
denote the \textit{desired} change of state (the desired momentum). Assume that we have an approximate feedforward
model of the inverse dynamics:
\begin{equation}\label{eq:uff}
\mathbf{u}_{ff} =
\mathbf{u}_{ff}(\mathbf{x},\mathbf{\dot{x}},\mathbf{v}(\mathbf{x}))
\end{equation}
which is an input-output  system receiving inputs (the state, the momentum, the desired momentum) and providing output
(the control vector $\mathbf{u}_{ff}:\mathbb{R}^{n \times n \times n} \rightarrow \mathbb{R}^p $). If this control
vector is used directly to influence the plant then it is called \textit{feedforward controller}. The perfect
feedforward control vector $\mathbf{u}^*_{ff}$ makes the plant to produce momentum $\mathbf{v}(\mathbf{x})$:
\begin{equation}\label{eq:perf_uff}
\mathbf{v}(\mathbf{x}) =
\mathbf{f}(\mathbf{x},\mathbf{u}^*_{ff}(\mathbf{x},\mathbf{\dot{x}},\mathbf{v}(\mathbf{x})))
\end{equation}
If the feedforward  control vector is imprecise then error (a difference between the desired momentum and the
experienced momentum) $\mathbf{e}_c=\mathbf{v}(\mathbf{x})-\mathbf{\dot{x}}$ appears. To correct this error, the (same
or another) model of the inverse dynamics can be used. This error correcting controller is called \textit{feedback
controller}\footnote{It is
to be noted that there is a reasonable freedom in the functional form of
the feedback and feedforward controllers.
\cite{szepesvari97approximate}.}: its inputs are the state, the momentum and the \textit{desired compensation}, i.e., $\mathbf{e}_c$. The output of the
feedback controller is subject to temporal integration. The output is called the feedback control vector $\mathbf{u}_{fb}$. The time
integrated and amplified output of this controller is used to correct the feedforward control vector:
\begin{eqnarray}\label{eq:u}
\mathbf{\dot{w}} = \Lambda \mathbf{u}_{fb} \\
\mathbf{u} = \mathbf{u}_{ff}+ \mathbf{w} \label{eq:u_}
\end{eqnarray}
i.e., $\mathbf{u} = \mathbf{u}_{ff}+ \Lambda \int \mathbf{u}_{fb} \,dt$ where $\Lambda$ denotes the (amplifying) gain
factor. Feedback vector $\mathbf{u}_{fb}$ disappears when $\mathbf{u}_{ff}=\mathbf{u}^*_{ff}$. Assume an approximate inverse dynamics of the following form:
\begin{equation}\label{eq:hat_phi}
\mathbf{u} = \mathbf{\hat{\Phi}}(\mathbf{x},\mathbf{\dot{x}}).
\end{equation}
A particular form of the feedback  control vector is simply a comparator that disappears
provided that
$\mathbf{u}_{ff}(\mathbf{x},\mathbf{\dot{x}},\mathbf{v}(\mathbf{x}))$
is perfect:
\begin{equation}\label{eq:ufb}
\mathbf{u}_{fb} =
\mathbf{\hat{\Phi}}(\mathbf{x},\mathbf{v}(\mathbf{x}))-\mathbf{\hat{\Phi}}(\mathbf{x},\mathbf{\dot{x}})
\end{equation}
This scheme is  depicted in Fig.~\ref{f:robust_controller}.

\subsection{From control architecture to reconstruction networks} \label{ss:simp_control}

\begin{figure}
\centering
 \includegraphics[width=76mm]{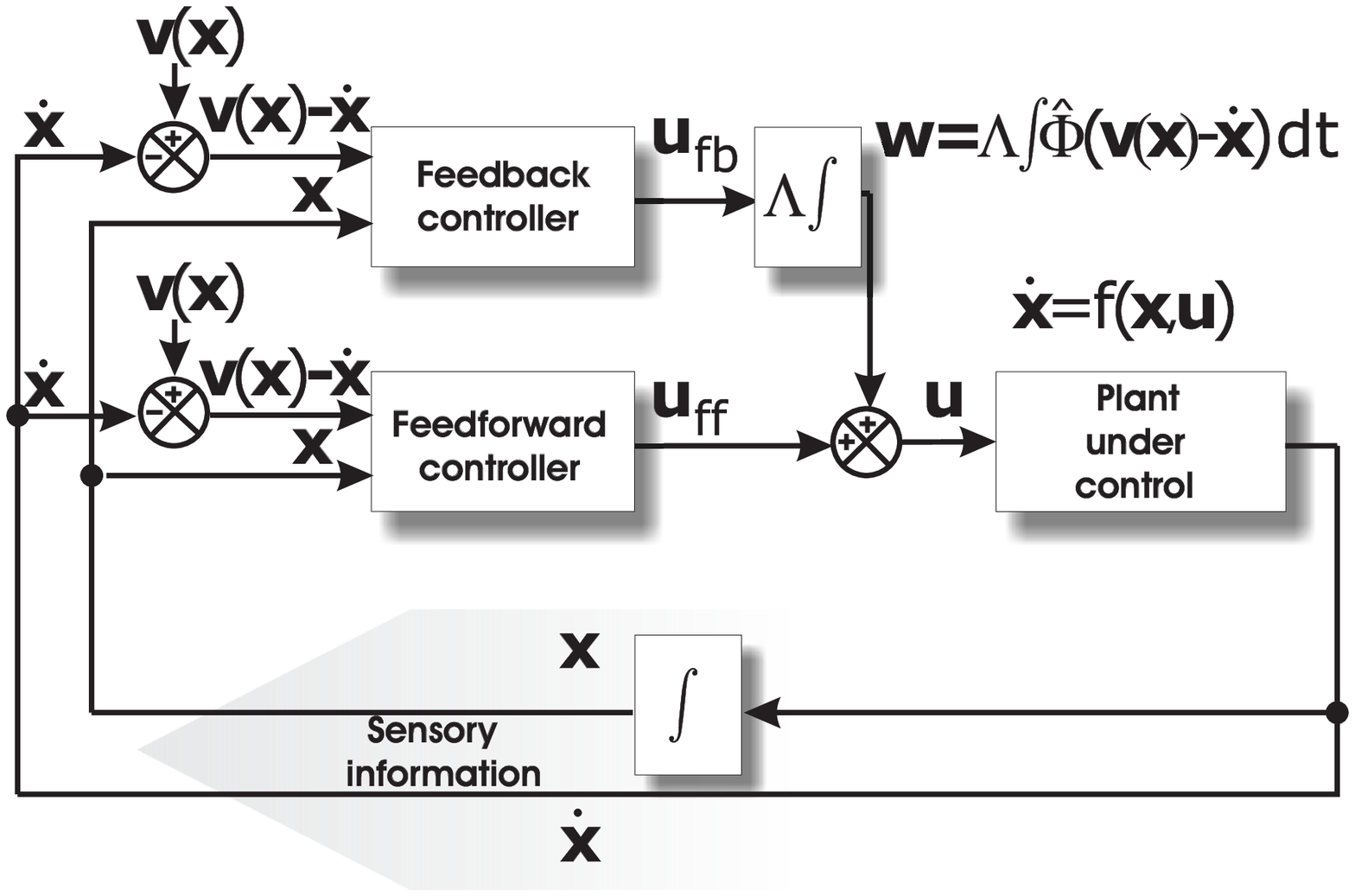}
  \caption{\textbf{Robust speed-field tracking controller using differences between desired and experienced quantities}\newline
  Feedforward and feedback controllers are the same, the output of
  the latter is integrated over time, amplified  and is added to
  the output of the former to control the plant. Both controllers make
  use of the simplified inverse dynamics of Fig.~\ref{f:robust_controller}(B)
  and both controller is inputted by the desired compensation
  $\mathbf{e}_c=\mathbf{v}(\mathbf{x})-\mathbf{\dot{x}}$. The scheme produces
  globally stable control under suitable conditions (Appendix \ref{s:appa}).
  }\label{f:simp_control}
\end{figure}

Our control architecture can be related to a
reconstruction network. To see this, first a particular (state
dependent but linear) form of the inverse dynamics is assumed
\cite{szepesvari97approximate}:
\begin{equation}\label{eq:hat_A_b}
\mathbf{\hat{\Phi}}(\mathbf{x},\mathbf{\dot{x}}) = \mathbf{\hat{A}}(\mathbf{x})\mathbf{\dot{x}} +
\mathbf{b}(\mathbf{x}).
\end{equation}
It has been shown that one can use the feedback controller in `feedforward position' without effecting stability properties \cite{szepesvari97approximate}:
\begin{equation}\label{eq:hat_A_b_1}
\mathbf{u}_{ff}=\mathbf{\hat{\Phi}}(\mathbf{x},\mathbf{v}(\mathbf{x}))-\mathbf{\hat{\Phi}}(\mathbf{x},\mathbf{\dot{x}})
\end{equation}
In this case the feedforward controller will never be perfect. Explicit modelling of $\mathbf{b}(\mathbf{x})$, is
unnecessary given that this quantity falls out in this
comparation-based controller
\begin{equation}\label{eq:hat_A_b_2}
\mathbf{u}_{fb}(\mathbf{x},\mathbf{\dot{x}},\mathbf{v}(\mathbf{x})) =
\mathbf{u}_{fb}(\mathbf{x},\mathbf{\dot{x}},\mathbf{v}(\mathbf{x})) = \mathbf{\hat{A}}(\mathbf{x})\left(
\mathbf{v}(\mathbf{x})- \mathbf{\dot{x}}\right).
\end{equation}
This simplified architecture, which is built of comparators is depicted in
Fig.~\ref{f:simp_control}. We note that (i) the control scheme is
capable of controlling plants of any order (Appendix
\ref{s:appb})
and (ii) it has attractive global stability properties
\cite{szepesvari97approximate}.

\begin{figure}[h!]
\centering
 \includegraphics[width=140mm]{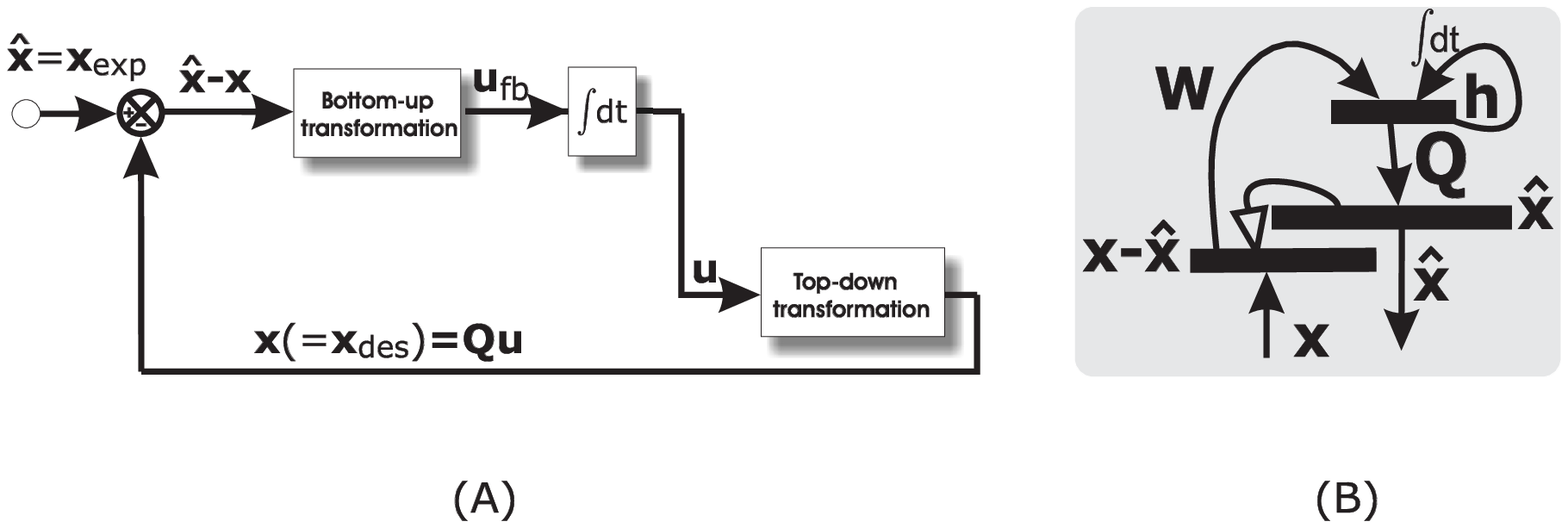}
  \caption{\textbf{Control architecture for first order plant and the equivalent reconstruction network}\newline
  \textbf{A:} Control architecture for first order plants and without feedforward controller.
  \newline For first order plant, the desired quantitiy is the desired state, whereas the experienced quantity
  is the experienced state. For the sake of comparison with subfigure \textbf{B}, desired state and experienced state are denoted
  by $\mathbf{x}$ and $\mathbf{\hat{x}}$, respectively.
  \newline \textbf{B:}  The corresponding reconstruction network.
  \newline Reconstruction network with the same dynamical properties.
  Input (vector) $\mathbf{x}$ is provided to the network. Input is \textit{compared}
  to `reconstructed input' (vector) $\mathbf{\hat{x}}$, which is generated by internal
  representation (vector) $\mathbf{h}$ through the `top-down' (TD) transformation
  (matrix) $\mathbf{Q}$. Mismatch (vector) $\mathbf{x}-\mathbf{\hat{x}}$ is delivered
  to correct the internal representation via `bottom-up' (BU) transformation (matrix)
  $\mathbf{W}$. Correction is achieved by temporal integration (i.e., adding up the
  correcting term and applying recurrent self-excitations) at the level of the internal representation. \textit{Note} the switch between
  experienced (sensed) and desired (to be matched) quantities.
}\label{f:control_and_recnet}
\end{figure}

Figure \ref{f:simp_control} can be further simplified (i) by
assuming a first order plant, i.e., a plant with dynamical
equation $\mathbf{\dot{x}} = \mathbf{f}(\mathbf{u})$ being
independent of the actual state  and (ii) by neglecting the
feedforward controller.\footnote{Restriction to first order plants
can be released  by the change of notations (Appendix
\ref{s:appa}). The control architecture can work without the
feedforward controller, but -- according to computational
experiments -- noise sensitivity increases considerably
\cite{szepesvari97approximate}.} Now, the experienced variable is
$\mathbf{x}$, whereas the desired variable is the desired state
denoted by $\mathbf{\hat{x}}$. The corresponding architecture is
shown in Fig.~\ref{f:control_and_recnet}(A). Figure
\ref{f:control_and_recnet}(B) depicts  a loop made of neural
network layers with the same dynamical properties (Appendix
\ref{s:appc}). From now on, let $\mathbf{W}$ and $\mathbf{Q}$
denote the `bottom-up' (BU) and the top-down (TD )connections of
the reconstruction network, respectively. There is, however, a
subtle difference between the two architectures: The control
network starts from a \textit{planned} desired state
$\mathbf{\hat{x}}$ and acts upon the plant to experience that
state. The reconstruction network experiences the state (i.e., the
input) $\mathbf{x}$ and acts upon a \textit{hidden layer} to
produce an the \textit{internal representation} $\mathbf{h}$ of
the network, which \textit{generates} a reconstructed input
$\mathbf{y}$ that should match the experienced input. The
reconstruction network is an auto-associator, equipped with a
hidden layer \cite{hinton83optimal,hinton94autoencoders}. The
reconstruction network is also a comparator that minimizes the
reconstruction error $\mathbf{e}=\mathbf{x}-\mathbf{\hat{x}}$. It
is worth noting that the sign of the difference is the opposite as
it is in the control architecture. The reason is that
reconstruction network follows the environment, whereas control
architecture manipulates it. The reconstruction network
\begin{enumerate}
    \item generates the reconstructed input $\mathbf{\hat{x}}$ via
    the `top-down' (TD) transformation, which is inputted by the
    hidden internal representation $\mathbf{h}$,
    \item \textit{compares} the input with the reconstructed input and
    produces the reconstruction error $\mathbf{e}=\mathbf{x}-\mathbf{\hat{x}}$,
    \item processes the reconstruction error via the `bottom-up' (BU)
    transformation and corrects the internal representation by
    that. (In continuous time, adding up `corrections' is equivalent to the temporal
    integration of the error.)
\end{enumerate}
Under certain  conditions -- matrix $\mathbf{WQ}$ of
Fig.~\ref{f:control_and_recnet}(B) should be positive definite
(see Appendix \ref{s:appc}) -- error compensation converges and
the network relaxes.
\begin{equation}\label{e:rec_simple}
\mathbf{\dot{h}}=\mathbf{W}(\mathbf{x}-\mathbf{Q}\mathbf{h})
\end{equation}

\subsection{Extended reconstruction network} \label{ss:elaboration}

The reconstruction network of Fig.~\ref{f:control_and_recnet}(B)
can be extended to fulfill particular constraints and
computational tasks. The extended network is depicted in
Fig.~\ref{f:Nonlin-KF}. The working of the network can be
understood as follows:

\begin{figure}
\centering
 \includegraphics[width=5cm]{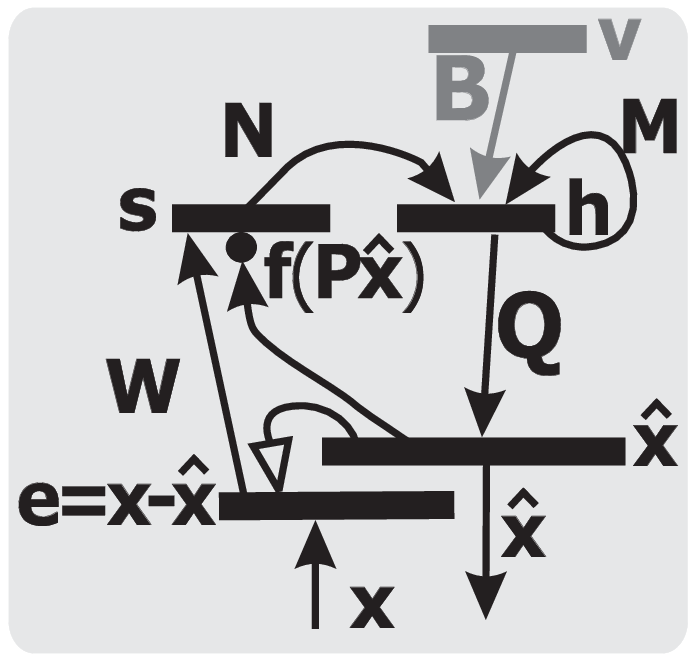}
  \caption{\textbf{Extended and controlled reconstruction network architecture}\newline
  Arrow with circle denote non-linear transformation.
  $\mathbf{x}$ and $\mathbf{\hat{x}}$: input and reconstructed input,
  $\mathbf{W}$ and $\mathbf{Q}$: bottom-up (BU) and top-down (TD) transformation,
  $\mathbf{s}$: BU processed reconstruction error with maximized information,
  $\mathbf{h}$: activity vector of the hidden (model) layer,
  $\mathbf{N}$: transformation from BU processed reconstruction error layer to the model
  layer,
  $\mathbf{P}=\mathbf{QN}^{-1}$: BU matrix of the inner loop,
  $f(\mathbf{P}\mathbf{\hat{x}})$ BU gate releasing vector, with sparsifying non-linearity
  $f(.)$,
  $\mathbf{B}$ and $\mathbf{v}$: tools of control. The reconstruction
  network can be considered as the environment of the control
  network, which provides the control vector $\mathbf{B}\mathbf{v}$. $\mathbf{v}$ is the desired momentum.
  Black: Reconstruction network architecture. Black and gray together: controlled reconstruction network.}\label{f:Nonlin-KF}
\end{figure}

Sensory  input vector $\mathbf{x}$ is compared to the
reconstructed input vector $\mathbf{\hat{x}}$. The error
$\mathbf{e}=\mathbf{x}-\mathbf{\hat{x}}$ is transformed by the
bottom-up (BU) transformation matrix $\mathbf{W}$ and forms the BU
transformed error $\mathbf{s}$. BU transformation maximizes BU
information transfer in order to facilitate reconstruction. BU
transformed error is passed to the internal representation layer
through transformation matrix $\mathbf{N}$ (the role of this
transformation shall be discussed later) and is added to the
internal representation $\mathbf{h}$ of the `hidden' (or model)
layer. The activity of the hidden layer is maintained by diagonal
elements of recurrent matrix $\mathbf{M}$. Considering the BU
error correction, matrix $\mathbf{M}$ can serve temporal
integration.

Beyond this temporal integration, off-diagonal elements of associative matrix $\mathbf{M}$ can perform temporal
prediction \cite{rao97dynamic,chrobak00physiological,poczos03arxivkalman}.

Reconstruction vector $\mathbf{\hat{x}}$ is generated by TD matrix
$\mathbf{Q}$. Note that any column of matrix $\mathbf{Q}$ could be
equal to (different) individual inputs. In this case,
reconstructed input will be the optimal linear combination of the
individual inputs. However, the learning principles of this TD
matrix can be more sophisticated than such a fast imprinting-like
encoding.

Our specific assumption on TD matrix $\mathbf{Q}$ is constrained
by our proposal on the resolution of the homunculus fallacy: the
function of the hidden layer should be spatio-temporal pattern
completion. Spatial pattern completion using prewired or
experienced correlations can complete missing pieces of
information. Representations using single (i.e., positive) sign
empower the learning of correlations. Algorithms capable of
finding positive components are called positive (non-negative)
matrix factorization algorithms
\cite{paatero94positive,lee99learning,xijin02learning}. Positive
matrix factorization together with pattern completion algorithms
may imply \cite{szatmary03robust} recognition by components
\cite{biederman87recognition}. That is, TD matrix $\mathbf{Q}$ is
assumed to accomplish positive matrix factorization. Matrix
$\mathbf{Q}$, which plays a major role in determining the relaxed
hidden activity, is considered the long-term memory of the
network.

As inner  loop is introduced in Fig.~\ref{f:Nonlin-KF} to perform
non-linear noise filtering or \textit{sparsification} of the BU
processed error. The sparsification matrix
$\mathbf{P}=(\mathbf{QN})^{-1}$ transforms reconstruction vector
$\mathbf{\hat{x}}$. Components of the output of this
transformation form the \textit{gate opening vector}. If a
component of this gate opening vector is below a certain threshold
then the corresponding component of the BU processed error is
diminished. There are theoretical works underpinning the idea: It
has been shown that wavelet denoising \cite{mallat98wavelet} can
be generalized to different databases using independent component
analysis (ICA)
\cite{jutten91blind,comon94independent,laheld94adaptive,bell95information,cardoso96equivalent,amari96new,karhunen97class,amari98natural}.
For a recent review on ICA, see \cite{hyvarinen99survey}. ICA
maximizes information transfer under the assumption that there are
hidden factors and the probability distribution of these factors
is equal to the product of the probability distribution of the
individual components. Estimations based on ICA components are
local: each component can be estimated separately.

Thresholding of independent components, alike to thresholding of
wavelet components decreases the structureless noise content of
the reconstructed input
\cite{hyvarinen99sparse,hyvarinen99sparse2}. The method is called
sparse code shrinkage (SCS). Relation between SCS and an
overcomplete reconstruction network with sparsifying non-linearity
\cite{OlFi97} has also been established
\cite{hyvarinen99sparse,hyvarinen99sparse2}.

Theoretical considerations
\cite{lorincz00parahippocampal,lorincz02mystery} indicate that (i)
matrices $\mathbf{P}$, $\mathbf{Q}$, $\mathbf{N}$, $\mathbf{W}$
can be tuned by Hebbian means. (ii) The reconstruction process
diminishes some of the components of the input by projecting into
to the subspace determined by the columns of matrix $\mathbf{Q}$.
(iii) Matrix $\mathbf{P}$ performs ICA on the reconstructed input
and thus denoising concerns the subspace of matrix $\mathbf{Q}$.
(iii) Matrix $\mathbf{W}$ performs ICA on the input and the two
ICA transformations may differ. (iv) Upon tuning, matrix
$\mathbf{QNW}$ becomes the identity matrix and (v) matrix
$\mathbf{P}$ becomes equal to matrix $\mathbf{W}$ and both perform
the same ICA transformation. (vi) The speed of tuning for matrices
$\mathbf{P}$, $\mathbf{Q}$ and $\mathbf{W}$ should be such that
tuning of TD matrix $\mathbf{Q}$ is the slowest and tuning of BU
matrix $\mathbf{W}$ is the fastest. We shall return to these
points later. In turn, this network does the following: (i) learns
to predict (near) future, (ii) maximizes BU information transfer,
and (iii) filters noise.

\subsection{Working of the extended architecture under control} \label{ss:KF_control}

\subsubsection{Working of the extended architecture} \label{ss:working}
The extended and controlled architecture satisfies the following non-linear equations:
\begin{equation}\label{e:controlled_rec_net}
  \mathbf{\dot{h}} = \underbrace{\mathbf{N}}_{(A)}f_{\mathbf{P}\mathbf{\hat{x}}}(\mathbf{W}\underbrace{(\mathbf{\hat{x}}-\mathbf{x})}_{(B)}
  + \underbrace{\mathbf{M}\mathbf{h}}_{(C)}  + \underbrace{\mathbf{B}\mathbf{v}}_{(D)},
\end{equation}
where BU error $\mathbf{e}=\mathbf{\hat{x}}-\mathbf{x}$ is the
mismatch between input and reconstructed input. This error vector
undergoes BU transformation (matrix $\mathbf{W}$) and forms the BU
error vector. Components of the BU error vector
(Eq.~\ref{e:controlled_rec_net}B) are subject to sparsification
(function $f_{\mathbf{P}\mathbf{\hat{x}}}(.)$) where matrix
$\mathbf{P}$ is determined by matrix $\mathbf{Q}$. Matrix
$\mathbf{M}$ (Eq.~\ref{e:controlled_rec_net}C) is responsible for
temporal integration, for prediction and for component based
completion of spatio-temporal patterns \cite{lorincz02mystery}.

\subsubsection{Control of the extended architecture} \label{ss:control_}

The reconstruction network of Fig.~\ref{f:Nonlin-KF} can be
controlled by acting on the internal representation. This is the
fourth term (D) of the r.h.s. of Eq.~\ref{e:controlled_rec_net}.
Control adds extra contribution, i.e., $\mathbf{B}\mathbf{v}$ to
the hidden layer. Vector $\mathbf{v}$ -- which can be a function
of the internal representation $\mathbf{h}$
($\mathbf{v}=\mathbf{v}(\mathbf{h})$) -- is the desired momentum.
From the point of view of the controller, either
$f(\mathbf{W}(\mathbf{\hat{x}}-\mathbf{x}))$ or its linear
transform $\mathbf{N}f(\mathbf{W}(\mathbf{\hat{x}}-\mathbf{x}))$
can serve as the experienced momentum. Clearly, there is no
warranty that the experienced momentum will match the desired one,
unless matrix $\mathbf{B}$ is properly tuned. For a learning
system, this match can not be warranted and thus a robust control
scheme that can enforce the desired quantities becomes a
necessity.

\subsection{Reconstruction network controlled by a robust controller} \label{ss:robust_KF}

\begin{figure}
\centering
 \includegraphics[width=140mm]{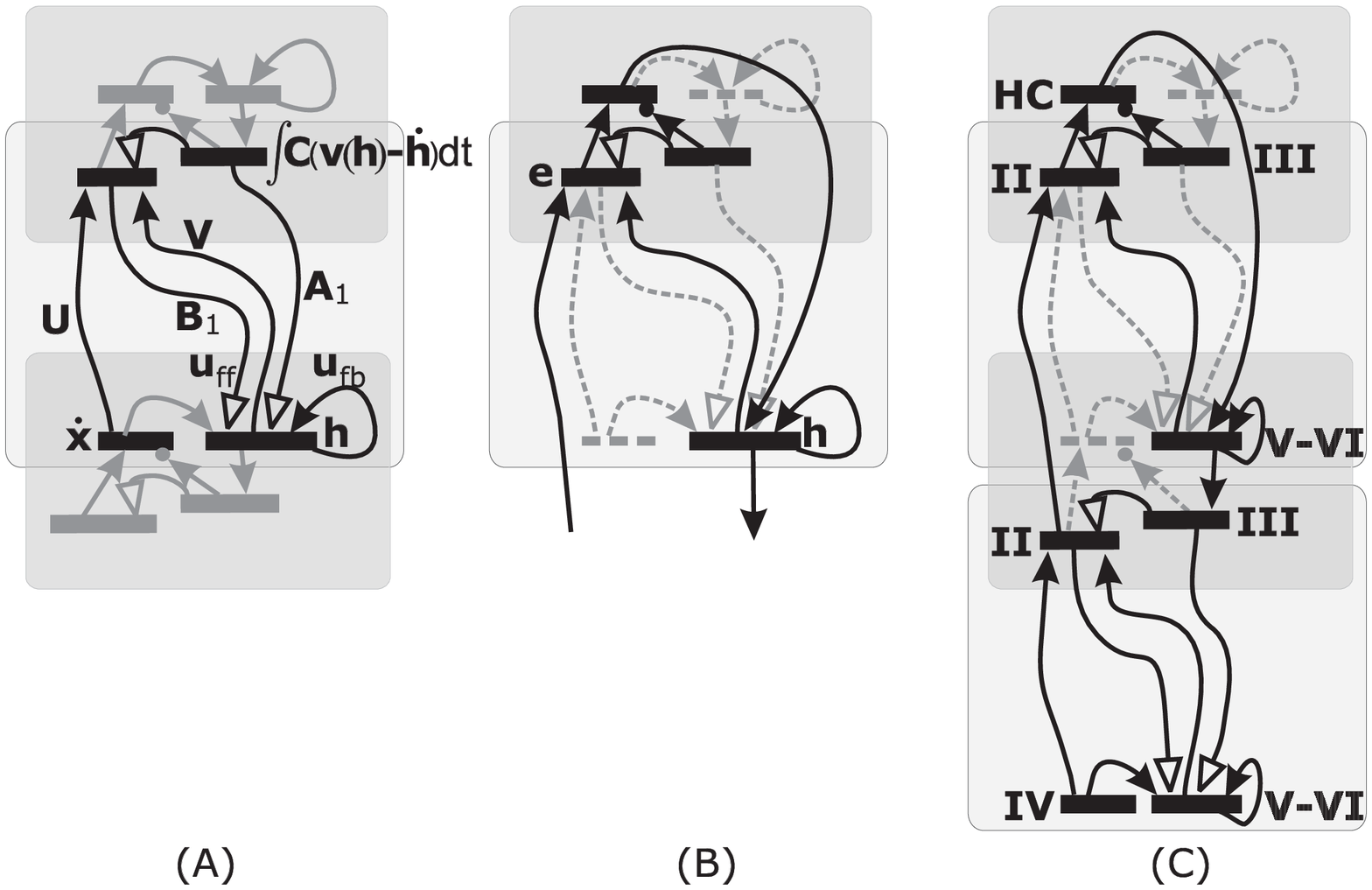}
  \caption{\textbf{Robust control of the extended architecture}\newline
  \textbf{A:} The two architectures, i.e., the reconstruction network and the
  robust control architecture are merged. Black color: components of the robust control architecture.
  Top and bottom: reconstruction networks. Matrix $\mathbf{U}$ carries $\mathbf{\dot{x}}$, the experienced
  speed. Also, $\mathbf{C}\mathbf{\dot{h}}= \mathbf{U}\mathbf{\dot{x}}$.
  Vector $\mathbf{V}\mathbf{h}$ may modulate the effect of matrix
  $\mathbf{U}$ and it can introduce state dependence.
  Matrices $\mathbf{A}_1$ and $\mathbf{B}_1$ are also components of the differencing controllers.
  $\mathbf{u}_{ff}=\mathbf{\hat{B}}(\mathbf{h})(\mathbf{v}(\mathbf{h})-\mathbf{\dot{h}})$.
 and $\mathbf{u}_{fb}=\int \mathbf{\hat{A}}(\mathbf{h})(\mathbf{v}(\mathbf{h})-\mathbf{\dot{h}})dt$ where
 matrices with hats comprise the effects of BU and TD transformations.
  (See text and Appendix \ref{s:appa}.) \newline \textbf{B:} The top of the hierarchy. Dashed gray
  arrows and dashed gray levels do not belong to the architecture at the top. The top plays double
  role: (1)It is a reconstruction network. (2) It is a robust controller, because mismatch influences
  the activities of a hidden layer at a lower level. \newline \textbf{C:} The top of the hierarchy
  together with a lower robust controller. Roman letters represent corresponding sublayers of areas of
  the neocortical hierarchy. Dark gray areas: robust control, light gray boxes: neocortical layers, HC: hippocampus.
 }\label{f:SDS_NC}
\end{figure}

The control scheme of Section \ref{ss:control} has several advantages (see Appendix \ref{s:appa}):
\begin{enumerate}
    \item As a result of the robustness, the learning of controlling is simplified: if the control action is
    \textit{sign proper}, i.e., the control action takes the system into
    the good direction, then control is ultimately uniformly bounded and globally stable. Moreover, the bound
    of the tracking-error can be made arbitrarily small.
    \item `Learning--by--doing' can be accomplished during controlling,
    no matter if sign properness is fulfilled or not: the actual state and momentum are to be associated with the actual control
    vector.
\end{enumerate}
Reconstruction  network extended by the robust controller is
depicted in Fig.~\ref{f:SDS_NC}(A) The working of  the
architecture on Fig.~\ref{f:SDS_NC} can be understood as follows:
First, let us consider the lower reconstruction network. This is
the `plant' (i.e., the `environment') to be controlled. The
controller is the upper reconstruction network. The lower network
passes its experienced BU error ($\mathbf{\dot{x}}$) to the upper
network through matrix $\mathbf{U}$. The effect of matrix
$\mathbf{U}$ may be modulated by vector $\mathbf{V} \mathbf{h}$.
The BU input to the upper network is $\mathbf{U}
\mathbf{\dot{x}}$, which -- up to a linear transformation -- is
equal to $\mathbf{\dot{h}}$. The desired momentum is provided by a
particular transformation (not shown explicitly in any of the
figures, but which is present in the neocortical structure
\cite{callaway99visual,calvin99columns}) that originates from the
internal representation and targets the corresponding
reconstruction error layer of the same reconstruction network. We
assume that this top-down signal is subtracted from the
experienced momentum. The difference of the desired and
experienced quantities forms the input to the feedforward
controller. The output of the feedforward controller is
$\mathbf{u}_{ff}=\mathbf{\hat{B}}(\mathbf{h})(\mathbf{v}(\mathbf{h})-\mathbf{\dot{h}})$.
Note that matrix $\mathbf{\hat{B}}$ contains both BU and TD
transformations. According to the working of the reconstruction
network, the input, i.e.,
$\mathbf{v}(\mathbf{h})-\mathbf{\dot{h}}$, undergoes SCS noise
filtering and temporal integration in the upper reconstruction
network to make the reconstructed input. Apart from a linear
transformation, this reconstructed input is the input to the
feedback controller (Eqs.~\ref{eq:u}, \ref{eq:u_} and
Fig.~\ref{f:simp_control}). The output of the feedback controller
is $\mathbf{u}_{fb}=\int
\mathbf{\hat{A}}(\mathbf{h})(\mathbf{v}(\mathbf{h})-\mathbf{\dot{h}})dt$,
where SCS noise filtering is not shown explicitly.

All components of the robust controller are now given and proper
operation can be achieved, provided that transformations are
sign-proper. In turn, the first and possibly the most problematic
step of the learning task is the shattering of the state space to
domains, within which the sign of control components does not
change. Whereas the finding of the sign-proper domains may be a
hard task, it is worth noting that there is no other condition
imposed on the BU and TD matrices. For example, these
transformations can be modulated vigorously, provided that
sign-properness is kept (Appendix \ref{s:appa}).

\subsection{Working of the perfectly tuned hierarchy} \label{ss:perfect}

Assume that all reconstruction networks are perfect: At all
levels, the products $\mathbf{QNW}$ of Fig.~\ref{f:Nonlin-KF} are
equal to $\mathbf{I}$, the identity transformation. Assume that
prediction is perfect, too. In this case, BU processing is error
free, no error appears, no error correction occurs and, in turn,
BU processing is as fast as in \textit{feedforward} networks.

\begin{figure}
\centering
 \includegraphics[width=140mm]{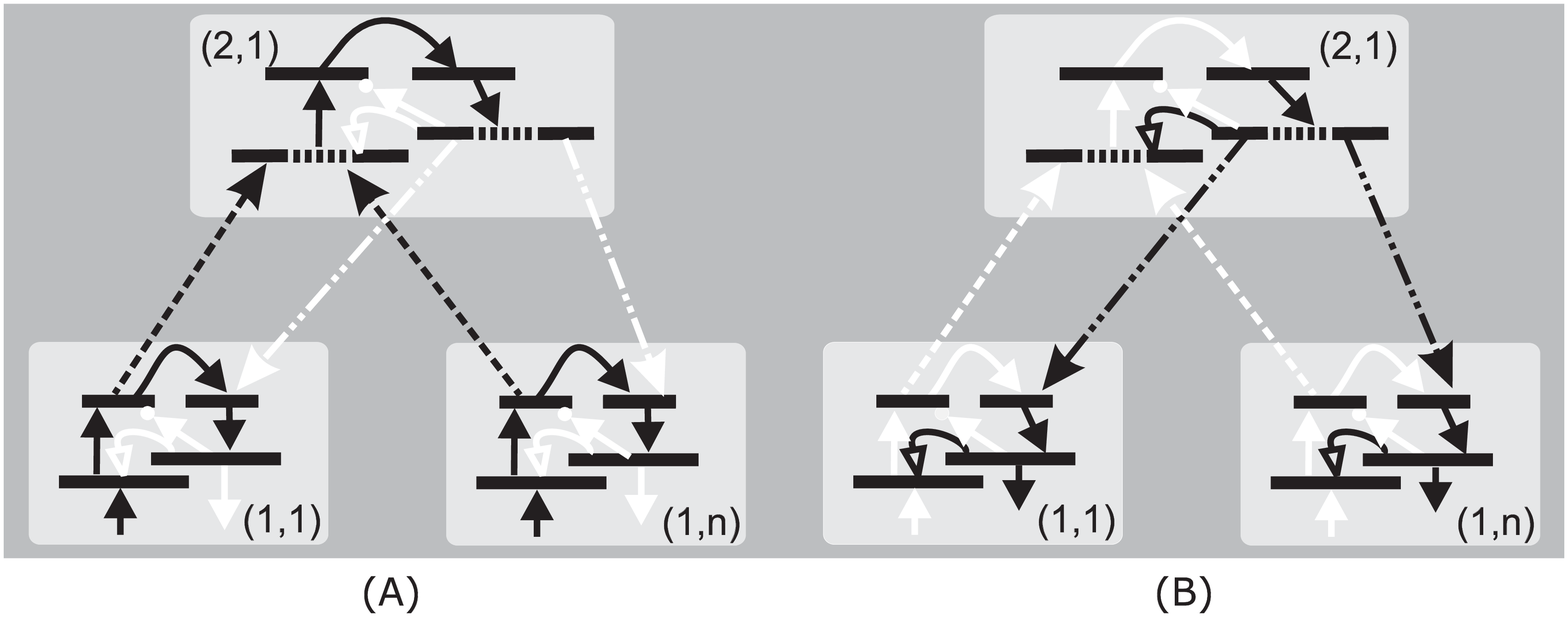}
  \caption{\textbf{Bottom-up filtering (A) and top-down control (B) are both feedforward for perfectly tuned
  networks}\newline Black arrows: the flow of information. White
  arrows: approximately silent connections.
  }\label{f:BU_TD}
\end{figure}

Similarly, for properly tuned inverse dynamics, top-down control
will not produce error and top-down processing works also as a
\textit{feedforward} control architecture. These features are
depicted for BU and TD processing in Fig.~\ref{f:BU_TD}(A) and
(B), respectively.

\subsection{Closing the loops of the hierarchy} \label{ss:closing}

We need to design  the top of the sensory processing hierarchy.
The top level receives information from different sensory systems,
or modalities. Pattern completion may reveal that some of the
components are missing or have been corrupted by noise and need to
be fixed. To have a coherent interpretation, the top may need to
correct such errors by influencing, i.e., controlling the internal
representations of lower layers, the `environment' of the top. In
turn, we need a twist at the top: Mismatch between input and
reconstructed input  at the top should be also the mismatch
between desired and experienced components. The two roles:
reconstruction and robust control should be merged at the top. The
solution is that the mismatch of the top reconstruction network is
also the mismatch of the inverse dynamics effecting the hidden
activity of a \textit{lower} reconstruction network. This twist is
shown in Fig.~\ref{f:SDS_NC}(B). Clearly, this is a reconstruction
network with hidden layer $\mathbf{h}$ displaced. Figure
\ref{f:SDS_NC}(C) depicts the twisted top together with a lower
control architecture. The hidden representation of the top
reconstruction network has the double role form the point of view
of its neighboring reconstruction error layers.

\begin{figure}
\centering
 \includegraphics[width=60mm]{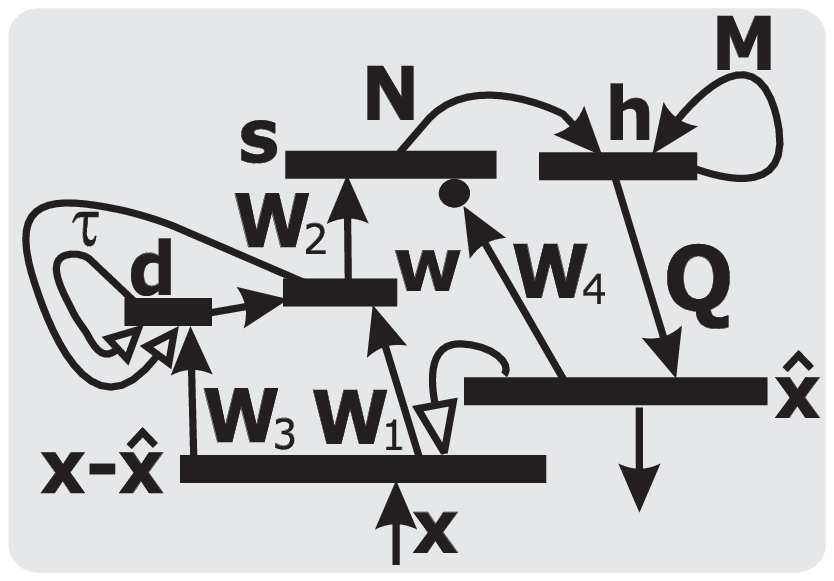}
  \caption{\textbf{The top of the hierarchy and its mapping.}\newline
  Maximization of BU information transfer is made in two steps: (i) whitening
  and (ii) ICA (separation). Blind source deconvolution sub-network has recurrent networks with
  long delays and removes temporal convolutions.
  \textit{Corresponding areas/layers} in the
  loop formed by the hippocampus and the entorhinal cortex (EC):
  (i) reconstruction error (vector $\mathbf{e}=\mathbf{\hat{x}}-\mathbf{x}$):
  layer II of EC, (ii) reconstructed input (vector $\mathbf{x}$): layer III of
  EC, (iii) hidden or model layer (vector $\mathbf{h}$): layers V and
  VI of EC, (iv) recurrent loop of hidden layer (matrix $\mathbf{M}$):
  associative structure of EC layers V and VI, (v) BU processing layer
  in computational order (whitened reconstruction error $\mathbf{w}$
  and  separated reconstruction error $\mathbf{s}$): area CA3 and
  area CA1 of the hippocampus, respectively, (vi) BU matrices $\mathbf{W}_1$
  and $\mathbf{W}_2$: EC afferents of the CA3 layer
  and Schaffer collaterals, respectively, (vii) recurrent loops with
  $\tau$ delays and vector $\mathbf{d}$: internal circuitry of the
  dentate gyrus and mossy cells of the dentate gyrus,
  respectively, (viii) matrix $\mathbf{W}_3$: perforant path
  afferents of the dentate gyrus, (ix) matrix $\mathbf{W}_4$:
  EC afferents of area CA1 of the hippocampus, (x) matrix $\mathbf{Q}$:
  deep layer to superficial layer connections of the EC, the long term memory
  components of the loop (xi) matrix $\mathbf{N}$: CA1 afferents
  of EC deep layers, the model layer of the loop. }\label{f:top_of_the_loops}
\end{figure}

This trick -- that is the double role architecture at the top of
the hierarchy -- turns bottom-up processing to top-down control.
However, there are additional constraints posed by information
maximization: As long as lower networks are not properly tuned,
processing is not feedforward and error correction gives rise to
temporally convolved signals (Appendix \ref{s:appc}). Temporal
convolution corrupts the maximization of information transfer. In
turn, temporal blind source deconvolution (BSD)
\cite{bell95information,torkkola96ablind,torkkola96bblind,lee97blind},
is necessary.

As it has been  emphasized by L{\H o}rincz and Buzs\'aki
\cite{lorincz00parahippocampal}, BSD in general, is very
demanding. The number of neurons and the number of connections
required by BSD are enormous. Fortunately, temporal convolution
induced by reconstruction networks has special properties and, in
principle, temporal deconvolution can be executed by a relatively
small -- but still `expensive' -- structure (see Appendix
\ref{s:appc} for details). This structure needs recurrent
connections with long and tunable delays. Given the robust control
properties of the hierarchy, approximately correct signals can be
enforced by starting from the top \cite{lorincz98forming}. Thus,
this expensive BSD structure is necessary only at the top of the
hierarchy. BSD at the top deconvolves the reconstruction error
before it is turned into a control signal. The reconstruction
network together with BSD structure made of tunable delay lines is
depicted in Fig.~\ref{f:top_of_the_loops}.

There is another difference between Figs.~\ref{f:Nonlin-KF} and
\ref{f:top_of_the_loops}: BU processing is executed in
Fig.~\ref{f:top_of_the_loops} in two layers. The two layers
perform two-step learning to maximize information transfer. The
transformation to the first layer whitens (i.e., it decorrelates
and normalizes components of the reconstruction error
\cite{laheld94adaptive}), whereas the transformation to the second
layer removes higher order correlations and, in turn, it develops
independent components. The two step learning algorithm is fast,
because it proceeds along the so called natural gradient (see,
e.g., \cite{amari98natural} and references therein). ICA can be
slower in the BU sparsification transformation if it follows the
one step learning rule of Bell and Sejnowski
\cite{bell95information}.
\begin{equation}\label{e:BS}
\Delta \mathbf{P} \propto f(\mathbf{s})\mathbf{\hat{x}}^T +
\mathbf{P}^{-1}
\end{equation}
where $f(.)$ denotes component-wise non-linearity. This learning
rule has two terms. One of them is a Hebbian term of the inputs
and the outputs of matrix $\mathbf{P}$. The other term is
proportional to matrix
\begin{equation}\label{e:BS_1}
\mathbf{P}^{-1}=\mathbf{NQ}
\end{equation}
that is to the rest of the loop. Matrix inversion, however, is not
necessary, the second term can be approximated, e.g., by noise
generated at the BU error layer and targeting the reconstructed
input layer. The reconstruction architecture warrants this
property through Eq.~\ref{e:BS_1}
\cite{lorincz00parahippocampal,lorincz02mystery}. In turn, the
order of learning is as follows: Novel structure not encoded into
the TD matrix but embedded in mismatch vector $\mathbf{e}$ is
blocked by SCS thresholding but undergoes fast ICA analysis and
develops high amplitude BU error components, which can not be
fully eliminated by thresholding. The access BU error undergoes
temporal integration at the level of the internal representation.
High activity components of vector $\mathbf{h}$ and high activity
components of mismatch vector $\mathbf{e}$ will induce Hebbian
learning in matrix $\mathbf{Q}$. This learning process is slow and
it is followed adiabatically (i.e., very closely) by matrix
$\mathbf{P}$. Given that learning of matrix $\mathbf{P}$ is
subject to Hebbian learning between outputs of matrix $\mathbf{W}$
and reconstruction error produced by matrix $\mathbf{Q}$, learning
of matrix $\mathbf{P}$ is kind of `supervised' by these matrices:
Upon TD matrix has incorporated the novel information, matrices
$\mathbf{W}$ and $\mathbf{P}$ become equal, provided that no novel
information has entered the loop. That is, we have the following
scenario: (a) novelty is blocked, (b) matrix $\mathbf{W}$ is
modified, (c) novelty is represented by a few large ICA components
(i.e., a few components of BU error increases, whereas many
components of BU error decreases), (d) large components overcome
sparsification, (e) TD long-term memory changes slowly, (f) this
slow change is closely followed by BU sparsification.

We note that a possible role of matrix $\mathbf{N}$ can be the
whitening of the output of the BU error layer that underwent
non-linear sparsification. This process, which advances Hebbian
learning for matrix $\mathbf{M}$, will be discussed elsewhere
\cite{szirtes03Kalman_arxiv}.

\section{Results: Straightforward mapping to sensory processing areas} \label{s:mapping}

Mapping -- based on the reconstruction network description -- has been thoroughly described elsewhere
\cite{lorincz00parahippocampal,chrobak00physiological,lorincz02mystery}. The control view \cite{lorincz98forming}
complements the basic structure of that mapping: beyond the function of connections \textit{between} neocortical
layers, it explains the function of connections of the neocortical layers that have not been modelled previously
\cite{lorincz02mystery}.

\subsection{Mapping to neocortical regions} \label{s:neocortex}

\begin{figure}
\centering
 \includegraphics[width=140mm]{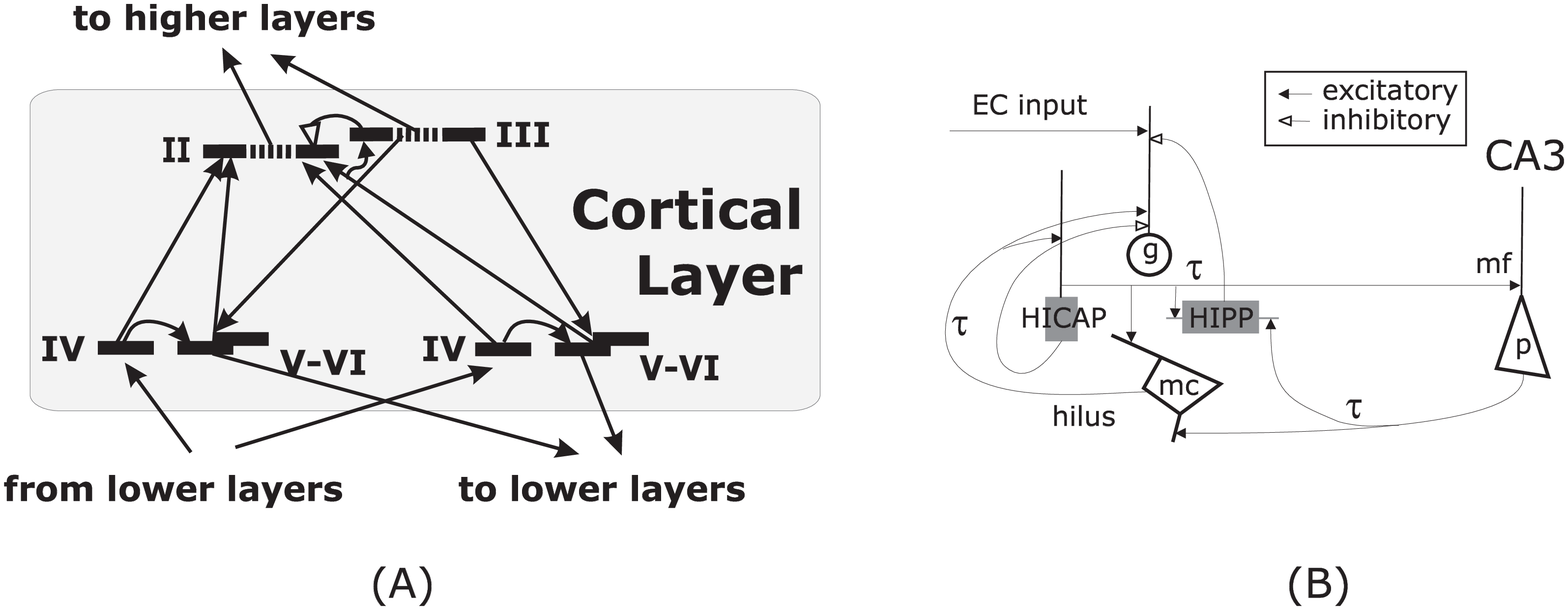}
  \caption{\textbf{Neocortical circuitry and the dentate gyrus}\newline
  \textbf{A:} Input: layer IV. Layer IV neurons send messages to layer II,
  layer III and layer VI.  Layer V and layer VI neurons receive
  messages from layer II and layer III. Layer V neurons provide
  feedback to layers II and III. Neurons of layers
  V and VI are connected (indirectly shown by the proximity of these
  layers). Neurons of layer II and III are
  also connected (only inhibitory connections are shown). Feedforward
  output to higher cortical layers: layers II and III. Feedback
  to lower layers: layer V.
\newline \textbf{B:} The excitatory connections between granule cells (g) and hilar mossy cells (mc) as well as in the g-CA3 pyramidal cells
  (p)-mc-g loop provide delay lines. Feedback inhibitory neurons innervate specific dendritic segments of the granule
  cells at the termination zone of the EC and mc afferents. HIPP: hilar interneuron with axonal termination in the
  perforant path zone; HICAP: hilar interneuron with axon termination in the commissural and association paths, mf: mossy fibers, $\tau$:
  synaptic delay. The activity of these interneurons also controls plastic changes (i. e., training) of the mc-gc synapses.  }\label{f:NC_from_book_and_dentate}
\end{figure}

The neocortex is made of six sub-layers
(Fig.~\ref{f:NC_from_book_and_dentate}(A)). The figure depicts the
most prominent connections between these sub-layers
\cite{lund88anatomical}. Input typically arrives at layer IV.
Layer IV neurons send messages to layer II and layer III (not
shown). Furthermore, layer IV neurons send messages also to layer
VI. Superficial neurons provide output down to layer V and VI.
There are connections between neurons of layer V and layer VI.
Neurons of layer II and III are also strongly connected. Layer V
provide feedback to layers II and III. The main output to higher
cortical layers emerges from layers II and III. The main feedback
to lower layers is provided by layer V. (For a review see, e.g.,
\cite{callaway99visual}.)

The theoretical model and the anatomical structure can be matched
only by assuming that reconstruction networks are laid
\textit{between} neocortical layers as it was denoted by the dark
gray areas in Fig.~\ref{f:SDS_NC}(A). On the other hand, robust
control is executed by the cortical layers, the light gray boxes
of Fig.~\ref{f:SDS_NC}(A). According to this figure, superficial
layers of the lower cortical layer \textit{and} deep layers of the
higher cortical layer form \textit{one} functional unit, the
reconstruction network. The reconstruction error and the
reconstructed input of this functional unit exert control action
on lower reconstruction networks. This controller is inputted by
the experienced state and by the BU processed reconstruction error
(i.e., the experienced momentum) of the lower reconstruction
network.

In this view, the V1 performs two functions: layers V and VI of
the primary visual cortex hold the internal representation of the
LGN, while layers II and III represent the input and the
reconstructed input of V2, respectively.

\subsection{Mapping to the hippocampal-entorhinal loop} \label{s:HCandNC}

The hippocampus (HC) is placed on the top of the hierarchy
(Fig.~\ref{f:SDS_NC}(C)). The HC incorporates a unique subunit,
the dentate gyrus, which, in our view, removes temporal
convolution of lower reconstruction networks. Loops required for
temporal deconvolution do exist in the dentate gyrus
(Fig.~\ref{f:NC_from_book_and_dentate}(B)). Notably, extreme long
delays (on the order of 500 ms), which is a crucial prediction of
our model \cite{lorincz00parahippocampal} has been found
experimentally \cite{henze02single}.

Besides temporal deconvolution, HC is also engaged in information
maximization and performs whitening and separation
(Fig.~\ref{f:top_of_the_loops}) performed by the CA3 and CA1
subfields of the hippocampus \footnote{No layer equivalent to the
IV$^{th}$ layer of the neocortex is present in this loop.}.
According to our model, the hippocampus plays two roles:
\begin{enumerate}
    \item HC acts upon the deep layers of the
    EC. In turn it can exert control action on the
    model layer of the EC.
    \item HC and EC together, form a reconstruction network, the
    EC-HC loop. This loop is special in that it may perform blind source
    deconvolution on its inputs.
\end{enumerate}
The two-phase  operation of the loop \cite{buzsaki89twostage}
ensures correct order of learning: analyzing and maximizing BU
information transfer and the encoding of top-down memory. Details
about the mapping to the EC-HC loop as well as details about
two-phase encoding can be found elsewhere
\cite{lorincz00parahippocampal}. According to that model, encoding
of long term memory is initiated by the recurrent collaterals (not
shown in the figure) of the CA3 area in one of the two-phases, the
sharp-wave phase.

\section{Discussion}
\label{s:disc}

\subsection{Relation to other models} \label{s:relations}

From the computational point of view, we should mention the model
of Gluck and Myers \cite{gluck93hippocampal}, which was designed
to perform reconstruction \textit{and} classification together for
modeling some properties of the hippocampus. This reconstruction
idea gains also importance in a recent model of Stainvas et al.
\cite{stainvas00improving}, in which reconstruction is used as
regularization constraint in classification task for creating
better representations.

Another approach has been proposed by Rao and Ballard. They have
put forth an integrating model
\cite{rao97dynamic,rao99optimal,rao99predictive} by exploring a
Kalman-filter analogy to cope with the input and system
uncertainties (treated as noise) and presented a hierarchy for
error correction and prediction using top-down inference from
higher levels. For spatial learning tasks, another Kalman-filter
based model has been proposed  with a biological mapping to the
hippocampus \cite{bousquet99is}. Kalman-filter is a kind of
reconstruction or generative network, which uses an internal
representation to generate expected inputs. Although the
Kalman-filter idea could be an efficient and plausible function
for sensory processing, the mapping of the proposed function onto
the anatomical, neurophysiological findings has not been
elaborated. Another problem is that noise filtering in
Kalman-filters requires matrix inversions. Also Kalman-filters
form loops, which are generally slow for the processing speeds
found in neocortical areas. Apparently, sensory processing in
neocortical areas is feedforward \cite{koch99predicting}.

\subsection{Algorithmic components of the CHF model} \label{s:algorithmic_comps}

The CHF has two basic components: (1) the reconstruction network
and (2) a robust control architecture.
\begin{enumerate}
    \item According to  Horn \cite{horn77understanding}, vision is inverse graphics. Our
    model generalizes this view by the old standing proposal that the hippocampus and its environment serve as a `comparator'
\cite{grastyan59hippocampal,sokolov63higher,vinogradova75functional}.
Auto-associators with hidden layers \cite{hinton94autoencoders},
i.e., reconstruction networks come to the sight by considering
these two suggestions together.
    \item The control aspect of sensory processing has been suggested long time ago: According to Diamond
\cite{diamond79subdivision} layer V could be seen as an extension
of the motor cortex in all areas, because layer V neurons send
inputs to basal ganglia, brain stem and sometimes even to the
spinal cord. This conjecture is reinforced by our mapping, where
layers V exert control actions on lower neocortical areas.
\end{enumerate}

The CHF model makes use of maximization of information transfer in
sensory processing, first suggested by Attneave
\cite{attneave54some} and Barlow \cite{barlow87learning}. We note
that ICA produces optimal representation for mean-field
approximation that empowers local (i.e., fast) approximation for
inferencing (information extraction from uncertain observations)
\cite{haft99model}.

One of the main algorithmic components of the CHF model, the so
called sparse code shrinkage (SCS) performs database optimized
denoising \cite{hyvarinen99sparse,hyvarinen99sparse2}. The
intriguing point is that denoising is experience based and, in
turn, novelty -- at first sight -- may appear as noise. The high
`noise content' blocked by sparsification is a direct sign of the
possibility of novel information. Novelty detection that precedes
recognition (the searching of the database) has been an old
mystery of information processing in the brain. The SCS algorithm
placed into the reconstruction network offers a solution here (see
later).

Speed-field tracking (SFT) based controlling is the other main
component of our model. Efforts have been made to derive the
entorhinal-hippocampal loop \cite{lorincz98forming} starting from
SFT. It can be motivated by
\begin{enumerate}
    \item the opinion that the medial paralimbic system (that includes the supplementary motor area as well as the
    anterior cingulate cortex, and that develops the elaborated basal ganglia thalamocortical loops) originates from the
    hippocampal cortex \cite{San69,Gold92} \textit{and} the strength of speed-field tracking in modelling
    those basal-ganglia thalamocortical loops \cite{Lorincz97IJNS,lorincz01ockham},
    \item the control aspect of layer V of neocortical areas \cite{diamond79subdivision},
    \item the view that the brain formulates signals that specify positions and directions of targets in extrapersonal
    space (see, e.g., \cite{zipser88backpropagation}) that resembles speed-field with dynamic path
    planning capacity \cite{FoRoSzLo96IJNS,SzLo98JRS},
    \item the mathematical tractability of speed-field tracking based control architecture \cite{szepesvari97neurocontroller,szepesvari97robust,szepesvari97approximate}.
\end{enumerate}

\subsection{Arguments for the CHF model} \label{ss:disc_attract}

\subsubsection{Experimental evidences} \label{ss:exp_evi}

As it has been  noted in the introduction, two predictions of the
model, (a) large and tunable temporal delaying capabilities of
neurons of the dentate gyrus and (b) persistent activities at deep
layers of the EC have been reinforced recently in
\cite{henze02single} and in \cite{egorov02graded}, respectively.
This latter prediction follows from the temporal integration at
the model layer, which are at the deep layers of the EC. In turn,
the CHF model has already passed two falsifying predictions.

\subsubsection{Implicit memory effects, order of
learning}\label{ss:implicit}

\subsubsection*{Order of learning:}
Learning steps  warranted by the model are as follows:
\begin{enumerate}
  \item The possibility of novelty is signaled by non-sparse BU processed error activity pattern.
  \item BU processed non-sparse activities, such as noise or novel information are gated by sparsification.
  \item `Noise' constantly undergoes fast maximization of information transfer in the BU processed error channel
  via ICA. Denoising (sparsification) cannot withhold high amplitude components of the BU processed error and
  information (i.e., structure embedded in noise) become available to the hidden layer of the model.
  \item Structured information is encoded into the top-down matrix, the long-term memory of our model.
\end{enumerate}
In turn, the architecture learns structure and rejects noise.

Some implicit memory effects emerge directly from the CHF model.

\subsubsection*{Repetition priming:}
The term  `priming' in a broader sense refers to the observation
that an earlier encounter with a given stimulus can modify
(`primes') the responding to the same or a related stimulus (see,
e.g., \cite{roediger93implicit} and references therein).

It has been shown by numerical simulations of reconstruction
networks using ICA for BU information maximization that repeated
presentations of not yet learned (novel or partially novel) inputs
shortens relaxation time of the reconstruction architecture
\textit{without} modifying LTM. The experienced decrease of
relaxation time has been interpreted as priming
\cite{lorincz02relating}. It can also be demonstrated that this
effect is enhanced by SCS denoising .

\subsubsection*{Repetition suppression and repetition enhancement:}
The neuronal  correlate of priming is thought to be
\textsl{repetition suppression}  \cite{desimone96neural}. This
belief is supported by the joint appearance of the two phenomena
in many experiments, see, e.g.,
\cite{buckner98functional,miller94parallel}: Both cognitive and
neurophysiological experiments show that neurons in the neocortex
respond with less and less activity in the case of repeated
stimuli (see e.g.,  \cite{schacter96conscious} and references
therein). This repetition decrement is often called `repetition
suppression' in the primate literature \cite{wiggs98properties}.
Numerical experiments demonstrate that repetition suppression
appears jointly with the shortening of relaxation time
\cite{lorincz02relating} in extended reconstruction networks. An
intriguing phenomenon is that during repetition suppression, a few
neurons do exhibit repetition enhancement \cite{desimone96neural};
an emergent property of our model: Repetition enhancement is shown
by those few units, which upon information maximization become
strongly activated and can break through the threshold of
sparsification \cite{lorincz02relating}.

\subsubsection*{Distributed nature of implicit memory effects:}
It is known that implicit memory effects, such as the recognition
of novelty, is distributed \cite{wan99different}. The order of
learning -- as described at the beginning of this subsection --
warrants that these effects, including novelty detection, may
occur at every reconstruction network and, in turn, it is
distributed in the CHF model. Similarly, the comparator function
is distributed in the model, too.

\subsubsection*{Prototype learning:}
Recent research  has provided evidence that category learning is
mediated by multiple neuronal systems in the brain (see, e.g.,
\cite{ashby01neurobiology} and references therein, but see
\cite{nosofsky98dissociation}). In contrast, when information is
accumulated from many exemplars and no verbal rules are easily
available, implicit mechanisms related to the basal ganglia may
operate. A good example for this latter type is a classic
prototype learning paradigm
\cite{posner68genesis,knowlton93learning,knowlton99what}.
Interestingly, prototype learning, which is spared in patients
with HC damage \cite{knowlton93learning} is impaired in Alzheimer
patients \cite{keri99classification,keri01are}. In reconstruction
networks, many exemplar based prototype learning may emerge in
different ways. We have also demonstrated that the adaptation of
the recurrent excitatory connections of the hidden layer
\cite{aszalos99generative,keri02categories} can explain the
impairment found in Alzheimer patients. Another candidate
structure is the recurrent excitatory connections of superficial
layers. These recurrent connections have not been included into
the CHF model yet. Their function will be conjectured later.

\subsubsection{Temporal compression}

The recurrent excitatory connections of the deep layers, which
perform spatio-temporal pattern completion, and the
error-correcting associative connections of the superficial
layers, together, fulfill the requirements of temporal compression
if network operation is not continuous but periodic
\cite{tsodyks96population}. Such temporal compression has been
observed in the hippocampus \cite{okeefe93phase,skaggs96theta}.

\subsubsection{Specific properties vs. generalization}

McClelland et al. \cite{mcclelland95why} emphasize the necessity
of a dual system for the seemingly contradictory tasks of learning
of specific properties and allowing for generalization. The CHF
model allows for an attractive solution here: Specific properties
can be encoded into the LTM, whereas generalization is allowed by
the flexible combination of LTM components at different levels of
the hierarchy by the control means. TD control, up to the limits
imposed by sign-properness, can distort, combine, and excite
memory components.

\subsubsection{Mathematical issues}\label{ss:math}

\subsubsection*{Adaptation and learning rules:}

As it has been described elsewhere
\cite{lorincz00parahippocampal,lorincz02mystery}, the loop
structure is advantageous for Hebbian learning. Neural activities
\textit{and} noise, together, with a relatively long (\~50 ms)
temporal window are required for whitening and the learning of
independent components in both BU pathways. Top-down matrices are
trained by the reconstruction error and the hidden layer activity
(the delta rule); the necessary signals for Hebbian learning are
available between the model layer and the reconstruction error
layer \cite{lorincz00parahippocampal,lorincz02mystery}.

The controller can learn by experimenting, i.e., by the
learning-by-doing scheme \cite{FoRoSzLo96IJNS}. No matter what the
desired parameters are, the experienced parameters need to be
associated to the control parameters to form an approximate
inverse dynamics. When the approximation becomes
\textit{sign-proper}, the control architecture will be
approximately precise \cite{lorincz01ockham}.

Adaptation of the robust controller is straightforward and occurs
via error feedback and temporal integration. It is an attractive
property of the control architecture that fast temporal changes of
different transformations of different networks of the hierarchy
(i.e., learning) can not disturb the stability of the controller
as long as the control remains sign-proper (Appendix \ref{s:appa})

\subsubsection*{Noise in the controller:}
It has been  noted in \cite{szepesvari97neurocontroller} that the
SDS scheme is sensitive to noise if the noise enters the system
just before the compensatory vector is integrated, i.e., if noise
affects $\dot{{\bf w}}=\mathbf{u}_{fb}$ of Eq.~\ref{eq:ufb}. Such
a noise can easily make the system unstable, because the
perturbation takes the form
\[
\Lambda \int_0^T n(t) dt,
\]
where $n(t)$ denotes the noise. Unfortunately, the boundedness of
integral of Eq.~\ref{eq:ufb} cannot be ensured for the general
case. Moreover, the amplitude of the perturbation will be
proportional to $\Lambda$. This means that increasing $\Lambda$
will also increase the perturbation of the system. This problem is
the problem of \textit{every} dynamic state feedback controller
provided that noise can enter precisely before the point where the
compensatory control signal is integrated through time. It is an
\textit{emerging} feature of the CHF architecture that this
problematic noise component can be diminished by optimized SCS
denoising.

On the other hand, the presence of noise is necessary and
advantageous in the CHF model. It is necessary for Hebbian
learning \cite{lorincz00parahippocampal} and it is also
advantageous to improve generalizing capability of ICA
\cite{hyvarinen00sparse}.

\subsubsection*{Reinforcement learning and behavioral relevance:}\label{ss:LTM}

It is a central issue if the robust controller can be incorporated
into the reinforcement learning (RL) framework or not. This
problem has been treated theoretically and answered positively:
The integration of robust controller and RL is possible within the
so called event learning framework, a novel form of reinforcement
learning
\cite{lorincz01event,szita02reinforcement,szita03epsilon,lorincz03event}.
In this scheme, time is broken into discrete time intervals and,
if temporal resolution is sufficient then, near optimal
performance with uncertain state descriptions can be achieved.
Some parameters that might change considerably (e.g., the mass
\cite{szepesvari97approximate} or the length of a robotic arm
\cite{lorincz01ockham}) may be left unobserved without affecting
near optimality, a rare feature in RL models. Furthermore, the
robust controller fits smoothly the reinforcement learning
interpretation of dopamine responses found in the basal ganglia
\cite{schultz97prediction,lorincz98basal}. Considering behavioral
relevance, there is no decision-making system embedded into the
CHF model; the CHF model is \textit{passive}. On the other hand,
we have succeeded to show that the CHF model, including the
reconstruction loop architecture and the robust controller can be
embedded into the reinforcement learning framework and near
optimal performance will be warranted. In turn, the model can be
easily extended by decision-making and planning.

In the CHF network, the hidden variables are controlled. It is
then another issue if a model with hidden variables can be
optimized using RL. We note that Kalman-filters, which are
tractable from the point of view of mathematical considerations,
can be seen as the mathematical approximation of the CHF scheme.
Kalman-filters have been integrated into the reinforcement
learning framework and convergence to the optimal solution is
warranted \cite{szita03kalman,szita03reinforcement}. To our best
knowledge, this is the first case when a partially observed
Markovian decision problem is shown to converge and to learn the
optimal policy. One emerging property of the derivation is that
the learning rule for the value of states has a Hebbian form
\cite{szita03kalman}. The weight factor of the learning rule is
proportional to the error of value prediction. Based on this
observation we may remark the followings:

\textit{Note (1)}  The CHF model, alike to its previous versions
\cite{lorincz98forming,lorincz00parahippocampal,lorincz02mystery}
is neither a model for episodic learning, nor a model for
incremental learning and does not fit such traditional
distinctions (see, e.g., \cite{gluck03computational}). On the one
hand, when information maximization is not modulated by behavioral
relevance, the CHF model is an incremental model. On the other
hand, the CHF model can be instantaneous, if behavioral relevance
(i.e., error in value estimation) increases learning efficiency.
Error in value estimation can modulate the Hebbian learning rule
of LTM and, in turn, input can be immediately encoded into the
reconstruction architecture.

\textit{Note (2)}  Long-term memory of the CHF model are not
permanent; they may change. Changes are subject to statistical
properties of the information (because information transfer is to
be maximized) and to behavioral relevance of the information. That
is, in the CHF model, `memory traces are unbound'
\cite{nader03memory}.

\subsection{Conjectures} \label{ss:disc_conjectures}

The CHF model allows us to make predictions. Some of the
conjectures qualify as attractive possibilities offered by the CHF
model, whereas others are falsifying predictions.

Principles of learning bottom-up and top-down transformations may
apply for the learning of the predictive matrix (i.e., matrix
$\mathbf{N}$) of the hidden layer, too. It is easy to show that
minimization of the square of the prediction error may lead to
Hebbian learning for this matrix and has the following form:
\begin{equation}\label{eq:delta_m}
\Delta \mathbf{M} \propto
\left(\mathbf{N\dot{h}}\right)\mathbf{h}^T
\end{equation}
provided that three conditions are met: (i) connections (synapses)
have access to the corresponding hidden layer activities, i.e., to
$\mathbf{h}$. (ii) The same connections have access to the error
of the hidden layer activities on the other side, i.e., to
$\mathbf{N\dot{h}}$ and (iii) input to the hidden layer is
whitened to avoid the necessity of multiplication by the inverse
of the correlation matrix $C(\mathbf{h},\mathbf{\dot{h}})$. From
condition (iii) it is conjectured that transformation
$\mathbf{N}$, which corrects the activity of the hidden layer,
whitens the sparsified BU error. This transformation has remained
unconstrained: The prescribed identity transformation of the full
loop can be achieved by rescaling matrices, e.g., the top-down
matrix. Whitening is also advantageous from the point of view of
learning: The learning rule of Eq.~\ref{eq:delta_m} follows the
natural gradient \cite{amari96new}. Condition (ii) implies that
one of the deep layers holds this whitened BU error, whereas the
other holds the hidden activities. Local circuits between the two
layers can provide the appropriate Hebbian training signals. Given
that neurons of layer V play control roles, it is layer VI, which
can hold the whitened BU error. If this is the case, indeed, then
sustained activity of neurons of layer V may be more pronounced
than that of neurons of layer VI.

The role of associative structures in the superficial layers is
not apparent in the CHF model. It is conjectured here that these
associative connections may serve error correction. We need to
tell error correction and pattern completion apart. Here, pattern
completion concern spatio-temporal patterns. Error correction, on
the other hand, is seen as a synchronous operation over actual
inputs, alike in content addressed memories \cite{grossberg89CAM}
or Hopfield network-like constructs \cite{wilson72excitatory}. The
dynamic Hopfield model \cite{kali00involvement} is a candidate CNS
model for layers with recurrent collaterals. Note that in the CHF
model \cite{lorincz00parahippocampal} the recurrent collateral
system of the CA3 subfield elicits randomly ordered temporal
replays of input sequences and serves to encode information into
the top-down memory and to diminish weak connections to improve
generalization capabilities \cite{hyvarinen00sparse}. This
replaying role is supported by experimental evidences
\cite{wilson94reactivation,skaggs96replay,nadasdy99replay,hirase01firing}.

Feedback connections between neocortical layers are generally more
numerous than the feedforward connections between neocortical
layers and these connections seem to have a weak functional role
(see, e.g., \cite{callaway99visual} and references therein). Also,
interaction within neocortical layers is much stronger than
feedback activities. On the other hand, these feedback connections
play a central role in the CHF model; these connections form the
long term memory of the architecture. In our view, this is an
apparent contradiction. The trick is that reconstruction is fast
and easy and, in the CHF model, it is feedforward in a well tuned
network. The hard problems are (i) how to choose amongst the
different possibilities, i.e., the `negotiation' between different
neocortical columns, different neocortical areas and different
sensory modalities. The possibility that different interpretations
may coexist in the brain has been made evident in the animal
experiments on binocular rivalry (see, e.g.,
\cite{leopold99multistable} and references therein) and in
experiments with several possible visual interpretations
\cite{leopold03visual}. (ii) How to express the decisions that
have been made? It is expected that on-going control signals -- no
matter if conscious or not -- are separated from background noise
by synchronous operation. Such synchronous signals dominate the
reconstruction process.

\textit{Falsifying prediction 1.}  An intriguing conjecture of the
CHF model is that disturbance of the feedback connections between
neocortical layers may corrupt apparent feedforward processing and
recognition but should not corrupt prototype learning.

\textit{Falsifying prediction 2.} The CHF model allows us to claim
that perceptual learning (see, e.g., \cite{schoups01practising}
and references therein) and categorical perception (see, e.g.,
\cite{goldstone94influences,harnad95learned} and references
therein) are two manifestations of the same long-term memory
effects. If so then (i) the place of encoding should include
almost exclusively the top-down LTM components of areas engaged in
sensory processing and (ii) frequent appearance without strong
behavioral relevance will barely modify the same components.
Nevertheless, modified inputs to higher layers may influence
bottom-up processing in areas above the place of encoding.

\section*{Acknowledgements}
\label{s:acknow}

This work was partially supported by Hungarian National Science
Foundation (Grant No. OTKA 32487). Special thanks are due to
Gy\"orgy Buzs\'aki for his enlightening and continuous support
during the long-course of our model construction. Careful reading
of the manuscript and helpful suggestions are gratefully
acknowledged to Gy\"orgy H\'ev\'izi and to G\'abor Szirtes.

\section{Appendices}
\label{s:app}


\subsection{Appendix A: Robustness and stable on-line adaptation}
\label{s:appa}

A plant is  called first order if its `position' (or configuration) determines the state of the plant \textit{and} if
the dynamical equation determines the momentum (of all parts of the plant). A plant is called second order if the state
is given by the configuration and the momentum \textit{and} if the dynamical equation determines the acceleration, and
so on. Higher order plants can be rewritten into the form of a set of first order differential equations by
concatenating `position', `momentum', etc. into the state vector (see also Appendix \ref{s:appb}).

Speed field  tracking (SFT) is not typical in the control literature, but arises naturally if we consider stationary
optimal-control problems such as path planning tasks \cite{HwaAh92}. Conventional control tasks, such as {\em
point-to-point control} and {\em trajectory tracking} cannot be exactly rewritten in the form of SFT and vice versa
\cite{szepesvari97neurocontroller,SzeLoNNW2}. SFT prescribes the speed vector $\mathbf{\dot{x}}$ of the plant as a
function of the state vector: \index{dynamic state-feedback}
\begin{equation}\label{eq:sft}
  \mathbf{\dot{x}} = \mathbf{v}(\mathbf{x})
\end{equation}
SFT task has  the advantage that the designer can incorporate several objectives into the form of the speed-field to be
tracked hence extend the model's range of possibilities.

The mathematical  treatment described here is a slight generalization of that of published in
\cite{szepesvari97approximate,szepesvari97neurocontroller}. The control scheme works for plants of any order, alike to
the original proof. The identity of the feedforward and the feedback controllers is released. This slight
generalization seems necessary to properly describe the superficial--to--deep layer wiring of the neocortex.

Let $D\subseteq \mathbf{R}^n$  denote the domain of the plant's state with the equation of motion given by
\begin{equation}\label{eq:plant}
  \mathbf{u} = \mathbf{B}(\mathbf{x})\mathbf{\dot{x}}+ \mathbf{b}(\mathbf{x})
\end{equation}
where $\mathbf{x}$  is the state vector of the plant and
$\mathbf{u} \in \mathbf{R}^m$ is the control. For simplicity the
dependence of $\mathbf{B}$ and $\mathbf{b}$ on $\mathbf{x}$ will
not be explicitly represented. Now let us assume that we have two
estimates of the true inverse-dynamics function
$\Phi(\mathbf{x},\mathbf{\dot{x}}) = \mathbf{B} \mathbf{\dot{x}} +
\mathbf{b}$, given by
$\mathbf{\hat{\Phi}}(\mathbf{x},\mathbf{\dot{x}})$ and
$\mathbf{\hat{\Psi}}(\mathbf{x},\mathbf{\dot{x}})$:
\begin{eqnarray}\label{eq:phi_psi}
\mathbf{\hat{\Phi}}(\mathbf{x},\mathbf{\dot{x}}) =
\mathbf{\hat{B}} \mathbf{\dot{x}} + \mathbf{\hat{b}}\\
\mathbf{\hat{\Psi}}(\mathbf{x},\mathbf{\dot{x}}) = \mathbf{\hat{A}} \mathbf{\dot{x}} + \mathbf{\hat{a}}
\end{eqnarray}

The SDS Feedback  Control equations \index{SDS feedback control} can then be written as
\begin{eqnarray}\label{eq:sdsb}
\mathbf{u} &=& \mathbf{u}_{ff}(\mathbf{x},\mathbf{\dot{x}},\mathbf{v}(\mathbf{x})) + \mathbf{w},\\
\mathbf{\dot{w}} & = & \Lambda \left( \hat{\Phi}( \mathbf{x}, \mathbf{v}(\mathbf{x}) ) - \hat{\Phi}( \mathbf{x},
\mathbf{\dot{x}})\right)
\end{eqnarray}
where $\mathbf{u}_{ff}$ is  the so called feedforward controller (to be specified later), $\Lambda>0$ is the gain of
feedback.

In what follows  the usual definition of positivity for fields of square matrices will be required. \vspace{0.5cm}

\noindent \textsc{Definition.}  \textit{Let $\mathbf{M}: D \rightarrow \mathbf{R}^{p\times p}$, $p>0$. $\mathbf{M}$ is
said to be positive definite uniformly over $D$ iff for all $\mathbf{x}\in D$ the term $\mathbf{M}(\mathbf{x})$ is
positive definite and there exists an $\epsilon>0$ such that $\lambda_{\min}(\mathbf{M(}\mathbf{x}))>\epsilon$ holds
for all $\mathbf{x}\in D$. Uniform negative definiteness can be similarly defined.}\vspace{0.5cm}

If $\mathbf{M}$ is a real  quadratic matrix then let $\mathbf{M}>0$ denote that $\mathbf{M}$ is positive definite.
Similarly, if $\mathbf{M}$ is a matrix field over $D$, let $\mathbf{M}>0$ denote that $\mathbf{M}$ is uniformly
positive definite over $D$ \index{uniformly positive definite matrix fields}.

\vspace{0.5cm} \noindent \textsc{Theorem.}  \textit{Assume that
the feedforward controller has the form
$$
\mathbf{u}_{ff}(\mathbf{x},\mathbf{\dot{x}},\mathbf{v}) = \mathbf{\hat{\Psi}}( \mathbf{x}, \mathbf{v} ) -
\mathbf{\hat{\Psi}}( \mathbf{x}, \mathbf{\dot{x}} ),
$$
which is similar as  the input of the feedback integrator.
Further, assume that the followings hold:}
\begin{enumerate}
\item $\mathbf{\hat{\Phi}}(\mathbf{x},\mathbf{\dot{x}}) = \mathbf{\hat{B}} \mathbf{\dot{x}} + \mathbf{\hat{b}}$ \item
$\mathbf{\hat{\Psi}}(\mathbf{x},\mathbf{\dot{x}}) = \mathbf{\hat{A}} \mathbf{\dot{x}} + \mathbf{\hat{a}}$ \textit{\item
$\mathbf{X}^T \mathbf{Y}$, where $\mathbf{X}$ and $\mathbf{Y} \in
\{\mathbf{A},\mathbf{\hat{A}},\mathbf{\hat{B}}\}$\footnote{By definition, $\mathbf{A}=\mathbf{B}$} are uniformly
positive definite over $D$} \textit{\item $\mathbf{A}$, $\mathbf{v}$, $\mathbf{b}$ are bounded and have uniformly
bounded derivatives w.r.t. $\mathbf{x}$ over $D$}
\end{enumerate}
\textit{Then for all $\Lambda>0$ the error of tracking $\mathbf{v}(\mathbf{x})$, $\mathbf{e} =
\mathbf{v}(\mathbf{x})-\mathbf{\dot{x}}$, is eventually uniformly bounded and, further, the eventual bound $b$ of the
tracking-error can be made arbitrarily small. More specifically $b=\mathcal{O}(1/\Lambda)$, and the eventual bound for
the time reaching $||\mathbf{e}||\leq b$ is proportional to $\Lambda$.}\vspace{0.5cm}

The proof of this theorem  relies on a Liapunov-function approach.
First of all, note that $\mathbf{\dot{w}} = \Lambda
\mathbf{\hat{B}}(\mathbf{v}-\mathbf{\dot{x}})$ and that
$\mathbf{u}=\mathbf{\hat{A}}(\mathbf{v}-\mathbf{\dot{x}})+\mathbf{w}$.
The relation $(\mathbf{B} + \mathbf{\hat{B}}) \mathbf{e} =
\mathbf{A} \mathbf{v} + \mathbf{b} - \mathbf{w}$ can be employed
to show that $L = {1\over 2} \mathbf{e}^T \left[ (\mathbf{A} +
\mathbf{\hat{A}})^T (\mathbf{A} + \mathbf{\hat{A}}) \right]
\mathbf{e}$ is an appropriate semi-Liapunov function. The proof
starts by differentiating $\mathbf{e}$ according to time:
\begin{eqnarray}\label{eq:diff_error}
  \frac{d}{dt} \mathbf{A} \mathbf{v} + \mathbf{b} - \mathbf{w} &=&
  \frac{\partial (\mathbf{A} \mathbf{v}+ \mathbf{b})}{\partial  \mathbf{x}}\mathbf{\dot{x}} -
  \mathbf{\dot{w}}\\
  &=&   \frac{\partial (\mathbf{A} \mathbf{v}+ \mathbf{b})}{\partial \mathbf{x}}\mathbf{\dot{x}} -
  \Lambda \mathbf{\hat{B}}(\mathbf{v}-\mathbf{\dot{x}})\\
  &=&  \mathbf{f}-\Lambda \mathbf{\hat{B}}\mathbf{e}
\end{eqnarray}
where $\frac{\partial  \mathbf{x}}{\partial \mathbf{x}}$ denotes the Jacobi derivative matrix of vector $\mathbf{x}$
according to vector $\mathbf{x}$ and $\mathbf{f}=\frac{\partial (\mathbf{A} \mathbf{v}+ \mathbf{b})}{\partial
\mathbf{x}}\mathbf{\dot{x}}$. In turn, the temporal derivative of our semi-Liapunov function assumes the following
form:
\begin{equation}\label{eq:errorb}
  \frac{d}{dt}L = \mathbf{e}^T
(\mathbf{A} + \mathbf{\hat{A}})^T \left[\mathbf{f}-\Lambda \mathbf{\hat{B}}\mathbf{e} \right]
\end{equation}
Then the proof can be completed by using the methods thoroughly
detailed in \cite{szepesvari97neurocontroller}. The proof compares
the two terms of the r.h.s. of Eq~\ref{eq:errorb} and notes that
the negative term, which can be made arbitrarily large, ensures
uniform boundedness.

The subtle point of  this theorem is the requirement that $\mathbf{A}^T \mathbf{\hat{A}}$ and $\mathbf{A}^T
\mathbf{\hat{B}}$ should be uniformly positive definite over $D$. The theorem gives rise to a global stability result.
Notice too that the particular form of the feedforward and feedback controllers make it unnecessary to build an
estimate of $\mathbf{b}$.

Note that in the  case of Eq.~\ref{eq:errorb} there is no dependence on the approximated inverse-dynamics. This fact
can be exploited to show that the above proof remains valid if $\mathbf{\hat{A}}$, $\mathbf{\hat{B}}$,
$\mathbf{\hat{a}}$ and $\mathbf{\hat{b}}$ vary in time but the conditions of the theorem remain valid at every instant.
Thus we get the following important corollary:\vspace{0.5cm}

\noindent \textsc{Corollary.}  \textit{Suppose that the conditions
of \textrm{Theorem} hold and also that
$\mathbf{\hat{A}}=\mathbf{\hat{A}}(t)$,
$\mathbf{\hat{B}}=\mathbf{\hat{B}}(t)$,
$\mathbf{\hat{a}}=\mathbf{\hat{a}}(t)$ and
$\mathbf{\hat{b}}=\mathbf{\hat{b}}(t)$. Next assume that
$\mathbf{B}^T \mathbf{\hat{B}}$ and $\mathbf{X}^T(t)
\mathbf{Y}(t)$, where $\mathbf{X}(t)$ and $\mathbf{Y}(t) \in
\{\mathbf{A}(t),\mathbf{\hat{A}}(t),\mathbf{\hat{B}}(t)\}$ are
uniformly positive-definite over $D$ and for all $t>0$, and that
$\mathbf{\hat{A}}(t)$ and $\mathbf{\hat{B}}(t)$ are bounded. Then
the conclusions of the above theorem still hold.}\vspace{0.5cm}

The uniform positive-definiteness conditions of the corollary
follow, e.g. when $\mathbf{\hat{B}}$ is bounded away from
singularities uniformly over $D$: an assumption often required in
adaptive control \cite{Sastry89}. It is clear, too that the
stability result does not depend on the specific adaptation
mechanism utilized, which is a fairly rare condition in adaptive
control theory. It also follows that gain $\Lambda$ can be adapted
during controlling.

However, one has to  provide an additional proof to show that the
conditions required for $\mathbf{\hat{A}}$ and $\mathbf{\hat{B}}$
are obeyed. If those conditions are not obeyed, then exponential
deviation may occur. In the case of exponential deviation, control
should be stopped before crash and and the `learning-by-doing'
procedure (Section \ref{s:disc}) can be invoked for the history
recently experienced.

\subsubsection*{Change of notations}

One may make use  of the following condensed notations: $\mathbf{x}_{des}=\mathbf{x}_{des}(\mathbf{x})$,
$\mathbf{x}_{exp}=\mathbf{x}_{exp}(\mathbf{x})$, where \textit{`des'} and \textit{`exp'} refer to desired and
experienced quantitites, which are dependent on the state $\mathbf{x}$. This notation can be augmented by the following
\textit{`subtraction rule'}:
\begin{equation}
\mathbf{x}_{des}-\mathbf{x}_{exp} = \left(%
\begin{array}{c}
  \mathbf{v}(\mathbf{x})-\mathbf{\dot{x}} \\
  \mathbf{x} \\
\end{array}%
\right)
\end{equation}
The notation and  the subtraction rule allows one to depict the robust control scheme of higher order plants (with no
feedforward controller) in the graphical form of Fig.~\ref{f:control_and_recnet}(A)

\subsection{Appendix B: Transcription of higher order differential
equations into first order differential equation} \label{s:appb}

Let us assume that  the dynamical equation of the plant has the following form:

\begin{equation}\label{eq:a1}
\frac{\partial ^n \mathbf{q}}{\partial t^n} = \mathbf{f}\left( \mathbf{q}, \frac{\partial \mathbf{q}}{\partial t} ,
\frac{\partial ^2 \mathbf{q}} {\partial t^2}, \ldots , \frac{\partial ^{n-1} \mathbf{q}}{\partial t^{n-1}} \right)
\end{equation}
and introduce the notations
\begin{eqnarray}\label{eq:a2}
  \mathbf{x} &=& (\mathbf{z}_1,\ldots ,\mathbf{z}_{n-1}) \\
  \mathbf{z}_1 &=& \mathbf{q}\\
  \mathbf{z}_{k} &=& \frac{\partial \mathbf{z}_{k-1}}{\partial t}  \,\,\,\, (k=2,\ldots , n-1)
\end{eqnarray}
then Eq.~\ref{eq:a1} can be rewritten as
\begin{equation}\label{eq:a3}
(\mathbf{\dot{x}}=)\,\, \frac{\partial \mathbf{x}}{\partial t^n} = \mathbf{g} \left( \mathbf{x}\right)
\end{equation}

\subsection{Appendix C: Relaxed deconvolving needs of signals
temporally convolved and mixed by reconstruction networks} \label{s:appc}

It is easy to show  by insertion that the dynamical equation
\begin{equation}\label{eq:h1}
\mathbf{\dot{h}}(t) = \mathbf{W}(\mathbf{x}-\mathbf{Q}\mathbf{h})
\end{equation}
of the reconstruction network of Fig.~\ref{f:control_and_recnet}(B) gives rise to the following solution
\begin{equation}\label{eq:h2}
\mathbf{h}(t) = \int_{-\infty}^t \exp\left(-\mathbf{WQ}(t-t')
 \right)\mathbf{W}\mathbf{x}(t')dt'
\end{equation}
The condition of  convergence is that $\mathbf{WQ}$ is positive definite.

Solution (\ref{eq:h2})  is forms a temporal convolution, which needs to be removed for proper maximization of
information transfer. Blind source deconvolution (BSD), in general, is demanding in terms of the number of neurons and
connectivity structure \cite{lee97blind}. However, the convolution of Eq.~\ref{eq:h2} can be simplified by mixing.
Diagonalizing matrix $\mathbf{WQ}$ as $\mathbf{WQ}=\mathbf{U}\mathbf{\Lambda}_{diag}\mathbf{U}^T$ with
$\mathbf{U}\mathbf{U}^T=\mathbf{U}^T\mathbf{U}=\mathbf{I}$, mixing $\mathbf{h}$ by $\mathbf{U}$, denoting the mixed
quantity by $\mathbf{\chi}$($=\mathbf{U}\mathbf{h}$) and introducing notations
$\mathbf{\xi}=\mathbf{U}\mathbf{W}\mathbf{x}$, one has
\begin{equation}\label{eq:h3}
\mathbf{\chi}(t) = \int_{-\infty}^t \exp\left(-\mathbf{\Lambda}_{diag}(t-t')
 \right)\mathbf{\xi}(t')dt'
\end{equation}
For BSD, Eq.~(\ref{eq:h3})  requires only diagonal delay lines and BSD is to be followed by a separate ICA
transformation, a much relaxed set of conditions. For some simulations, and for arguments on convergence-divergence
patterns in the dentate gyrus, see \cite{lorincz00parahippocampal}.

\bibliographystyle{amsplain}
\providecommand{\bysame}{\leavevmode\hbox to3em{\hrulefill}\thinspace}
\providecommand{\MR}{\relax\ifhmode\unskip\space\fi MR }
\providecommand{\MRhref}[2]{%
  \href{http://www.ams.org/mathscinet-getitem?mr=#1}{#2}
} \providecommand{\href}[2]{#2}

\end{document}